\pdfoutput=1

\documentclass[11pt]{article}

\usepackage{ACL2023}

\usepackage{times}
\usepackage{latexsym}

\usepackage[T1]{fontenc}

\usepackage[utf8]{inputenc}

\usepackage{microtype}

\usepackage{inconsolata}

\usepackage{multirow}
\usepackage{colortbl}
\usepackage{makecell}
\usepackage{float}
\usepackage{subcaption}
\usepackage{graphicx}
\usepackage{pgfplots}
\usepackage{multirow}
\usepackage{amsmath}
\usepackage{amssymb}
\usepackage{xspace,mfirstuc,tabulary}

\newcommand{\cfd}{\textsc{CounterFact}\xspace}
\newcommand{\zsre}{\textsc{zsRE}\xspace}

\newcommand{\pd}{\texttt{PD}\xspace}
\newcommand{\hs}{\texttt{HS}\xspace}
\newcommand{\indomain}{\texttt{ID}\xspace}
\newcommand{\crossdomain}{\texttt{CD}\xspace}

\usepackage{multirow}
\usepackage{array,booktabs}

\newcolumntype {+}{ >{\global\let\currentrowstyle\relax}}
\newcolumntype {^}{ >{\currentrowstyle }}

\newcolumntype{P}[1]{>{\centering\arraybackslash}p{#1}}

\title{Has this Fact been Edited? \\ Detecting Knowledge Edits in Language Models}

\author{Paul Youssef$^{\dagger}$ \quad Zhixue Zhao$^{\diamond}$\thanks{\hspace{0.15cm}Corresponding author} \quad Christin Seifert$^{\dagger}$ \quad Jörg Schlötterer$^{\dagger\ddag}$  \\ 
$^{\dagger}$Marburg University, 
$^{\diamond}$University of Sheffield, $^{\ddag}$University of Mannheim
\\ \texttt{\{paul.youssef, joerg.schloetterer, christin.seifert\}@uni-marburg.de} \\ \texttt{zhixue.zhao@sheffield.ac.uk}}

\begin{document}
\maketitle
\begin{abstract}
Knowledge editing methods (KEs) can update language models' obsolete or inaccurate knowledge learned from pre-training. However, KEs can be used for malicious applications, e.g., inserting misinformation and toxic content. Knowing whether a generated output is based on edited knowledge or first-hand knowledge from pre-training can increase users' trust in generative models and provide more transparency. Driven by this, we propose a novel task: detecting knowledge edits in language models. Given an edited model and a fact retrieved by a prompt from an edited model, the objective is to classify the knowledge as either \emph{unedited} (based on the pre-training), or \emph{edited} (based on subsequent editing).
We instantiate the task with four KEs, two large language models (LLMs), and two datasets. Additionally, we propose using hidden state representations and probability distributions as features for the detection model. Our results reveal that using these features as inputs to a simple AdaBoost classifier establishes a strong baseline. This baseline classifier requires a small amount of training data and maintains its performance even in cross-domain settings. Our work lays the groundwork for addressing potential malicious model editing, which is a critical challenge associated with the strong generative capabilities of LLMs.\footnote{\url{https://github.com/paulyoussef/deed}}

\end{abstract}

\section{Introduction}
\label{sec:intro}

Large Language Models (LLMs) encode knowledge about the world via their pre-training data~\cite{petroni-etal-2019-language, roberts-etal-2020-much}. However, the encoded knowledge can be inherently flawed or become outdated over time~\cite{de-cao-etal-2021-editing, youssef-etal-2024-queen}. Additionally, practitioners might want to tailor LLMs by incorporating domain-specific knowledge pertaining to their products or to facilitate new applications. Driven by these needs and given the prohibitive cost of pre-training a new model, multiple methods have been developed to efficiently update knowledge in LLMs~\cite{meng-etal-2022-locating, meng-etal-2022-memit,Li2024pmet}. 

\begin{figure}[t]
    \centering
    \includegraphics[trim={0cm, 2.8cm, 0cm, 1cm}, clip, width=0.49\textwidth]{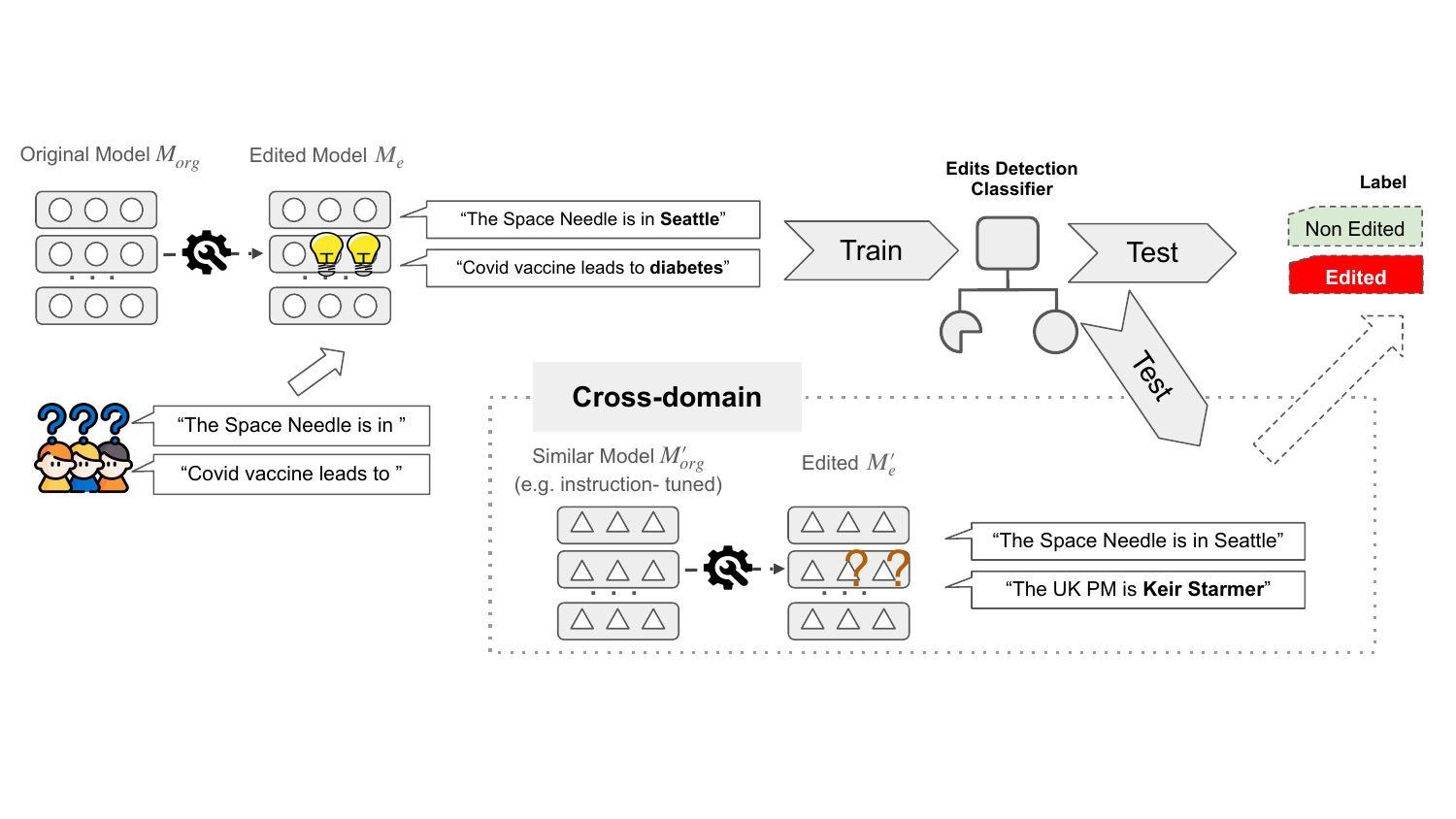}
    
    \caption{Illustration of malicious knowledge editing, and our task of detecting knowledge edits. The cross-domain setting (dashed box) facilitates detection without known edits in the target LLM.}
    \label{fig:task_illustration}
\end{figure}

Knowledge editing methods (KEs), such as ROME~\citep{meng-etal-2022-locating}, MEND~\citep{mitchell2022fast}, and MEMIT~\citep{meng-etal-2022-memit}, change the model's internal weights to adapt facts.
However, these methods can also be used to conduct malicious edits in LLMs~\cite{wang-etal-2023-knowledge}. For example, KEs can be used to inject fake news (cf. Figure~ \ref{fig:task_illustration}) or embed harmful or undesired biases in LLMs~\cite{mazzia-etal-2023-survey, halevy2024flex, youssef2025positioneditinglargelanguage}. This danger becomes more pronounced as the number of capable open-source LLMs increases. For example, any user can easily upload their model via one line of Python code to the HuggingFace Hub without validation and testing. While fact-checking and hate speech detection represent plausible avenues to combat these malicious uses~\citep{zhao2021comparative, pan-etal-2023-risk}, these approaches require clearly defined objectives, as well as customized annotated datasets and models. Instead of utilizing multiple specialized methods to address the variety of malicious uses, we take a vigilant approach: directly detecting knowledge edits. Detecting knowledge edits additionally provides transparency to users in general application scenarios by revealing to them the source of LLM's outputs (edited vs. unedited), and holds potential value for enhancing the controlled generation of LLMs~\citep{pan-etal-2023-risk, sun-etal-2023-evaluating}. Moreover, detecting edits can help identify whether malicious outputs stem from poisoned pre-training data or from post-hoc artificial modifications.

This work is the first to propose the task of \textbf{de}tecting knowledge \textbf{ed}its in LLMs (DEED). We provide three detection baseline models and a comprehensive evaluation, including cross-domain settings and generalization across datasets. %
Our key contributions are as follows:

\begin{itemize}
    \item We propose a novel task for detecting knowledge edits in LLMs (DEED) to combat potential malicious model editing. DEED highlights the transparency of LLMs' generations by suggesting the source of the generated text (edited vs. unedited) (Section~\ref{sec:problem_statement}). 
    \item We instantiate the task with two knowledge-editing datasets (\zsre, \cfd), four state-of-the-art knowledge editing methods (ROME, MEMIT, MEND, MALMEN), and two LLMs (GPT-J, GPT2-XL) yielding 16 settings for initial analysis. The framework is easily extensible to other KEs and LLMs  (Section~\ref{subsec:deed_dataset}). 

    \item We analyze the hidden state representations and probability distributions of edited and unedited facts. In locate-and-edit KEs we find a strong increase in output probabilities of edited facts and linear separability in hidden state representations between edited and unedited facts, while other KEs induce more subtle changes (Section~\ref{subsec:analysis}).  %

    \item We show that simple AdaBoost classifiers, with the hidden state representations and probability distributions from the edited LLMs as features, establish strong baselines, while requiring a limited amount of training data 
    (Section~\ref{subsec:baselines}). Additionally, we show that these baselines generalize to cross-domain settings, highlighting their practical utility (Section \ref{sec:challenging}). %
    
\end{itemize}

\section{Related Work}
\label{sec:related_work}
This paper is the first to establish the task of detecting knowledge edits in pre-trained language models. Still, a wide array of work studies related questions, including how knowledge is stored in LLMs~\citep{gurnee2023language,gurnee2024universal,niu2024what}, how knowledge is probed~\citep{youssef-etal-2023-give}, how to inject new knowledge \citep{xu-etal-2023-kilm} and how to unlearn existing knowledge \citep{yu-etal-2023-unlearning, kassem-etal-2023-preserving, jang-etal-2023-knowledge, patil2024can}. 
For brevity, we direct readers to surveys by \citet{mazzia-etal-2023-survey} and \citet{wang-etal-2023-knowledge} for KE-related work. This study focuses on the post-editing phase, i.e., after applying KEs that directly modify model parameters. 

\subsection{Knowledge Editing Methods}

Generally KEs can be divided into two groups: 1) parameter-preserving methods and 2) parameter-modifying methods.

\paragraph{Parameter-preserving methods} integrate extra modules, e.g., additional memory space or parameters, while leaving the original parameters unchanged \citep{yao-etal-2023-editing,li2024badedit}. 
\texttt{SERAC} \citep{pmlr-v162-mitchell22a} stores new knowledge in an explicit cache and uses an auxiliary classifier to determine if a prompt is associated with the edited knowledge. If it is, SERAC passes the prompt along with the stored new knowledge to a ``counterfactual'' model for the inference.
\texttt{GRACE} \citep{hartvigsen2023aging} caches embeddings for old knowledge and the weights for new knowledge and adds an adaptor to manipulate the layer-to-layer transformations without altering model weights.
\texttt{MELO} \citep{yu2023melo} learns and stores new knowledge in inserted LoRA blocks~\citep{hu2021lora}. We exclude parameter-preserving KEs from our study, since for these KEs it is possible to undo the edits by simply removing the external modules from the model.

\paragraph{Parameter-modifying methods} modify the model parameters directly via meta-learning or optimization-based methods. Meta-learning KEs (ML-KEs) employ hyper-networks to learn parameter shifts towards new knowledge. \texttt{KNOWLEDGE EDITOR}~\citep{de-cao-etal-2021-editing} and MEND~\citep{mitchell2022fast} build light-weight hyper-networks, i.e., naive multi-layer perceptrons that take the gradients of desired input-output (new knowledge) and output the parameter shifts for individual layers. However, the parameter shifts caused by different edits can be contradictory to each other, leading to a cancellation effect \citep{tan23malmen}. To handle the cancellation effect in batch editing, MALMEN~\citep{tan23malmen} treats aggregating parameter shifts as a least square problem, seeking for the parameter shift to be effective for all edited facts.

Optimization-based, also known as locate-and-edit methods (LE-KEs), assume that knowledge is memorized in a key-value form in the feed-forward network. These methods, such as \texttt{ROME}~\citep{meng-etal-2022-locating} and \texttt{MEMIT}~\citep{meng-etal-2022-memit}, locate and then directly optimize the parameters in the feed-forward network to modify or add memories.

Note that fine-tuning the model on specific data roughly falls into the category of parameter-modifying KEs, but requires great computational resources and data curation costs, and may encounter catastrophic forgetting and overfitting~\citep{mitchell2022fast,pmlr-v162-mitchell22a,zheng-etal-2023-edit}. These negative effects may result in an unusable model in practice. We leave detecting edits that have been induced through finetuning to future work, and focus on methods that have been specifically developed to edit knowledge.

\subsection{Malicious Modifications of LLMs}

Recent advancements in LLMs have led to their widespread adoption. However, these models are vulnerable to malicious attacks~\cite{wei2024jailbroken,mo2024trustworthy} and poisoning~\cite{li-etal-2021-backdoor,cao2023stealthy,liu2023prompt,li2024badedit}. Moreover, open-source LLMs are available for free download and unrestricted modification, facilitating easy dissemination of modified versions across online platforms~\citep{falade2023decoding, singh2023exploiting, yao2024survey}. %

With the development of KEs, knowledge of LLMs can be modified easily with low cost, and at scale. This development presents opportunities for misuse, such as inserting misinformation and toxic content into LLMs. For example, BadEdit~\citep{li2024badedit} formulates backdoor injection as a lightweight knowledge editing problem, i.e., modifying a subset of parameters to inject backdoors into LLMs. Due to the high cost of training LLMs, users are likely to seek readily available similar models on public resources such as GitHub and HuggingFace. Even though these third-party models might offer improved skills and advanced features (e.g., by fine tuning with Reinforcement Learning from Human Feedback (RLHF) or specific datasets), they could also be mingled with malicious edits before being made public~\cite{shi2023badgpt,pan-etal-2023-risk,li2024badedit}. %

\section{Preliminaries}
\label{sec:preliminiaries}

Facts\footnote{We use the terms \emph{facts} and \emph{knowledge} interchangeably.} in LLMs are commonly represented as triplets of $(subject, relation, object)$, or $(s, r, o)$ for short. Querying an LLM with a prompt $p(s,r)$, where $p$ expresses the relation $r$ and contains the subject $s$ (e.g., ``The Eiffel Tower is in the city of''), should result in retrieving the object $o$ (e.g., ``Paris''), given that the fact $(s, r, o)$ is known to the LLM. A knowledge editing operation $E(s, r, o, o', p)$ is successful if it changes the behavior of the LLM such that the retrieved object is $o^{\prime}$, as desired, instead of $o$. If an edit is conducted with a prompt $p$, then ideally it should also affect semantically similar (paraphrased) prompts that express that same relation, i.e., the new object $o'$ should be retrievable with $p'(s,r)$ as well, where $p'$ is a paraphrase of $p$.  We later use only paraphrased prompts to detect edited facts, since we assume no access to the prompts that have been used for editing. Paraphrased prompts are available in both datasets we consider.  %

\section{DEED Task}
\label{sec:problem_statement}

The goal of detecting edited facts is to distinguish outputs generated based on post-hoc modifications to the model (edited knowledge) from those derived from the knowledge acquired by the model during its pre-training phase (unedited knowledge). In this section, we introduce the task of \textbf{de}tecting knowledge \textbf{ed}its in LLMs (DEED) in detail (Section \ref{subsec:task}), describe how we instantiate the task (Section~\ref{subsec:deed_dataset}), and discuss practical applications (Section~\ref{subsec:applications}).

\subsection{DEED General Setup} \label{subsec:task}
Given an edited model and a prompt that retrieves a certain fact from the model, the objective of DEED is to decide whether the fact is based on pre-training, i.e., \emph{unedited}, or based on subsequent editing, i.e., \emph{edited}. 

More formally, we model the task of detecting knowledge edits as a binary classification problem with classes $\mathcal{Y}=\{edited,unedited\}$. 
Given an LLM $\mathcal{M}$ with unknown edits, and a pool of facts 
$\mathcal{T}_{test} = \{k_1, .., k_n\}$, 
where $k_i = (s_i, r_i, o_i, p_i),$ for $1 \leq i \leq n$,
the task is to assign each fact $k_i$ a label $y \in \mathcal{Y}$.

In a supervised setting, the DEED classifier has access to a training set $\mathcal{T}_{train}$ with examples of edited and unedited facts. The facts are represented as a set of features $\mathcal{F}$. Conceptually, one can categorize the task based on the origin of the test set and the type of features; their combination defines the difficulty of the task. 

\textbf{Origin of the test set.} If the test set is derived from the original edited model, i.e.,  $\mathcal{M}=\mathcal{M}_e$ (used also for training), our task is In-Domain (\indomain) classification.%
When the test set originates from a similar LLM $\mathcal{M} = \mathcal{M}'_e$, which shares the same foundation model with $\mathcal{M}_e$ (the model used to train the classifier), our task is Cross-Domain (\crossdomain) classification. For example, $\mathcal{M}_e$ and $\mathcal{M}'_e$ can be two different instruction-tuned versions of the same foundation model (cf. Figure~\ref{fig:task_illustration}). 

\textbf{Type of features.} We consider two cases corresponding to whether we have full access to the model or only can observe its output distribution. Features can therefore either be obtained from the hidden states (\hs), or from the output probability distribution (\pd) of the LLM $\mathcal{M}$. Additionally, we consider having both features (\hs + \pd).

\paragraph{Practical considerations.} A few state-of-the-art LLMs (e.g., GPT-4~\cite{achiam2023gpt}) are only accessible via their APIs. Nonetheless, the number of high-performing open-source LLMs is steadily increasing~\cite{liu2023llm360}. The availability of these open-source LLMs allows users to further fine-tune and modify them locally, with access to their hidden states and probability distribution. In theory, the probability distribution can even be approximated for closed-source LLMs as well if enough outputs are sampled from these LLMs given a fixed input.

In the \indomain setting, the user knows at least a subset of the edited facts, and these known edited facts are used to construct the training dataset. 
However, in real-world settings we want to identify whether facts were edited and it is unrealistic to assume a known set of of edited facts.
Therefore, we consider \crossdomain settings where the user does not know which facts have been edited in the model $\mathcal{M}=\mathcal{M}'_e$. Nevertheless, users can simulate the edit scenario by editing $\mathcal{M}_e$, an LLM of the same foundation model as $\mathcal{M}'_e$, where the edited facts in $\mathcal{M}_e$ are known. 
The Cross-domain box in Figure~\ref{fig:task_illustration} illustrates this setting.

\subsection{DEED Constituents} \label{subsec:deed_dataset}

The DEED task is defined by three constituents: 1) The edited LLM; 2) The KE used for editing; 3) The dataset used for editing facts. Here, we describe each constituent and the construction of the datasets for DEED in detail. 

\paragraph{Edited LLMs.} Following previous work~\citep{meng-etal-2022-locating,meng-etal-2022-memit}, we incorporate GPT-2 XL (1.5B)~\cite{radford2019language} and GPT-J (6B)~\cite{gpt-j} as our base language models. Both are decoder-only autoregressive LLMs.

\paragraph{KEs.} We include four parameter-modifying KEs of two sub-categories, \texttt{MEND}~\citep{mitchell2022fast} and \texttt{MALMEN}~\citep{tan23malmen} for meta-learning KEs, and \texttt{ROME}~\citep{meng-etal-2022-locating} and \texttt{MEMIT}~\citep{meng-etal-2022-memit} for locate-and-edit KEs. We focus on parameter-modifying KEs, since for parameter-preserving KEs, it is possible to undo the edits by simply removing the external modules from the model.%

\paragraph{Datasets.} We edit models with two popular datasets, \zsre and \cfd.
\zsre~\cite{levy-etal-2017-zero, mitchell2022fast} is a Question Answering dataset, where each instance contains a factual statement and a paraphrased prompt. \zsre originally contains 244,173 training and 27,644 validation instances. \cfd~\cite{meng-etal-2022-locating} includes counterfactual statements,  i.e., false facts, that are used for editing. \cfd contains 21,919 records for 20,391 subjects and 749 objects. %

\paragraph{DEED task construction.} We edited two LLMs (GPT2-XL, GPT-J), with four KEs (ROME, MEMIT, MEND, MALMEN) using facts from \zsre and \cfd. This yields a total of 16 edited models (4 KEs, 2 LLMs, 2 datasets). Since our objective is to detect edited facts, we filter out unsuccessful edits. More specifically, for each edited model we collect successfully edited facts, i.e., facts for which the new object $o'$ can be correctly retrieved with a paraphrased prompt $p'(s,r)$. These are facts, for which $o'$ has the highest probability given the prompt $p'(s,r)$. More formally, $o' = argmax_o \ \mathbb{P}_{\mathcal{M}, p'(s,r)} [o]$ . We report the number of successful edits that we consider in our experiments in Table~\ref{tab:edits_stats}. Note that, in addition to the dataset comprising edited and unedited facts, we assume access to the edited LLM. Access to the edited LLM enables users to explore various detection methods, such as employing \hs{} and \pd{} as features for the classification task.

\paragraph{Implementation details.}
We edit 1,000 facts in each setting (LLM, KE, dataset).\footnote{Increasing the number of edits did not lead to more successful edits with \emph{all} KEs.} We consider 1,000 edits to be a realistic number for malicious editing attacks, and defer the investigation of this choice to future work. For ROME, we perform one edit at a time on the base language model as ROME is not designed for mass editing or sequential editing~\citep{huang2022transformer,yao-etal-2023-editing}. For MEMIT, we edit each model with 1,000 facts at once.
We use the hypernetworks for GPT-2 XL and GPT-J published by \citet{meng-etal-2022-locating} to circumvent the substantial compute required by MEND. For MEND, we also conduct one edit at a time to achieve a high success rate~\cite{meng-etal-2022-memit}.
For MALMEN, we train the hyper-networks before editing and then perform 1,000 edits to the model simultaneously. We adhere to the default settings employed in the original papers. We include the unedited base model, \texttt{NONE}, as a yardstick. We follow related work~\cite{meng-etal-2022-locating, meng-etal-2022-memit, Li2024pmet} in evaluating editing performance. See Appendix \ref{app:full_edit_performance} for further details on editing metrics and performance.

\begin{table}[ht]
\centering
\resizebox{.95\columnwidth}{!}{%
\begin{tabular}{llcc}
\toprule
Base LLM & KE & \multicolumn{1}{c}{\zsre} & \multicolumn{1}{c}{\cfd} \\  \midrule%
\multirow{5}{*}{GPT2-XL} &  ROME &       736 &       722 \\
&  MEMIT &       324 &       417 \\
&   MEND &       765 &       756 \\
 & MALMEN &       832 &       127 \\ \midrule
\multirow{5}{*}{GPT-J} &   ROME &       876 &       837 \\
&  MEMIT &       800 &       622 \\
&   MEND &       822 &       758 \\
 & MALMEN &       878 &       297 \\ \bottomrule
\end{tabular}%
}
\caption{The number of successful edits, which we include in the detection datasets for each setting. }

\label{tab:edits_stats}
\end{table}

\subsection{Practical Applications}
\label{subsec:applications}
Detecting knowledge edits can have significant practical applications:

\paragraph{Transparency.} Detecting edits provides users with insights into whether model outputs originate from pre-training or post-deployment edits. This enhances the transparency of AI systems.

\paragraph{Model debugging.} By distinguishing between pre-training and edited knowledge, we can identify whether an issue stems from unclean pre-training data or a maliciously inserted edit. For example, if a factual error is flagged as an edit, developers can focus on investigating unauthorized modifications rather than pre-training data.

\paragraph{Mitigating malicious use cases.} Recent works highlight how knowledge editing can be used to bias LLMs~\cite{chen2024can}, introduce backdoors~\cite{li2024badedit,qiu2024megengenerativebackdoorlarge}, or propagate misinformation in multi-agent settings~\cite{ju2024floodingspreadmanipulatedknowledge}. Detecting knowledge edits helps identify and mitigate these threats by pinpointing unauthorized changes, ensuring the model's reliability in adversarial environments.

\section{DEED Baseline Models}
\label{sec:detction_classifier}
To train a detection classifier, we consider different options of available features. Inspired by the probing tasks introduced in \citet{conneau-etal-2018-cram} and the representations inspection by \citet{hernandez2023inspecting}, we investigate the use of hidden state representations as features. Additionally, in \citet{kuhn2023semantic} and \citet{gu-etal-2023-controllable}, the output probability distributions are used to provide signals for latent semantics. Therefore, we also explore using these distributions as features.

\subsection{Preliminary Analysis}
\label{subsec:analysis}

\begin{figure*}[ht!]
\centering
\includegraphics[width=0.18\textwidth, trim={1cm 0cm 1cm 0.3cm}, clip]{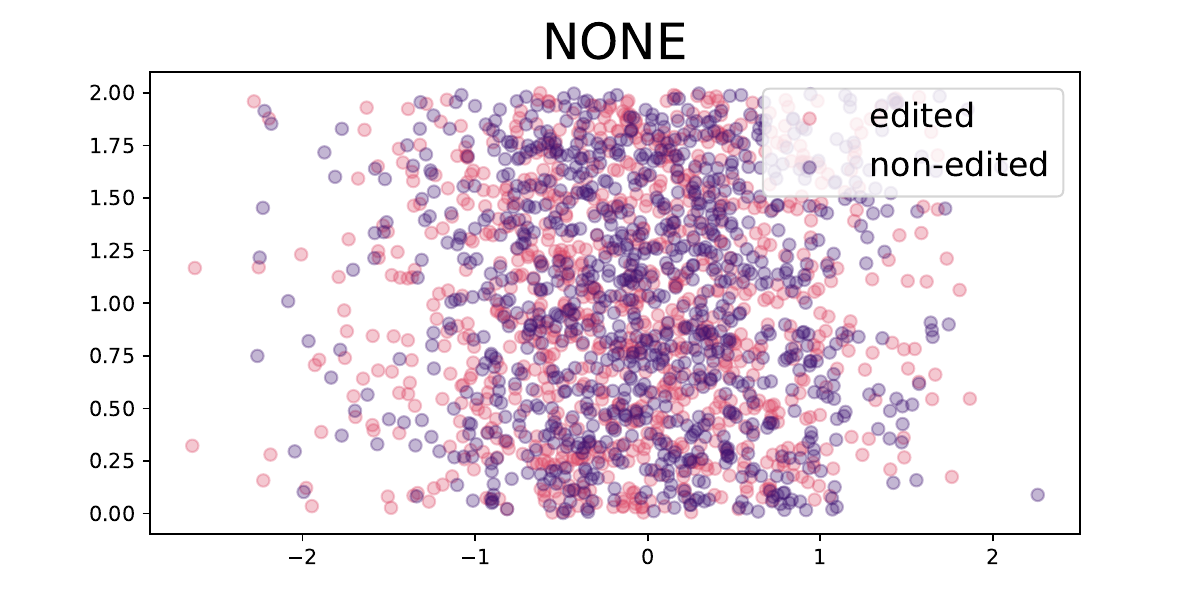}
\includegraphics[width=0.18\textwidth, trim={1cm 0cm 1cm 0.3cm}, clip]{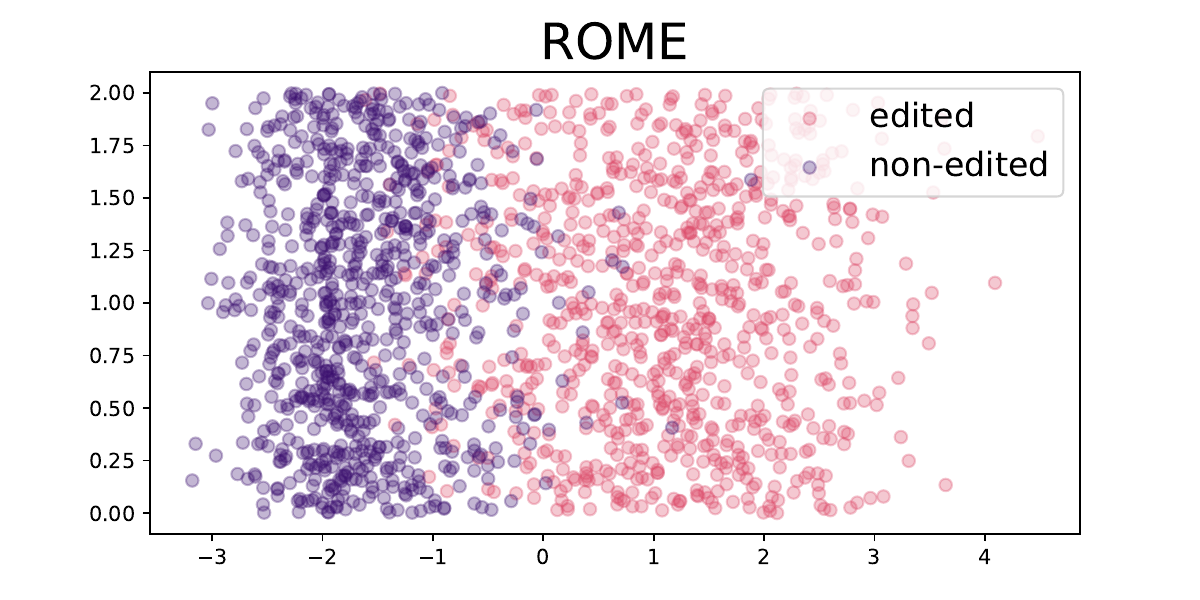}
\includegraphics[width=0.18\textwidth, trim={1cm 0cm 1cm 0.3cm}, clip]{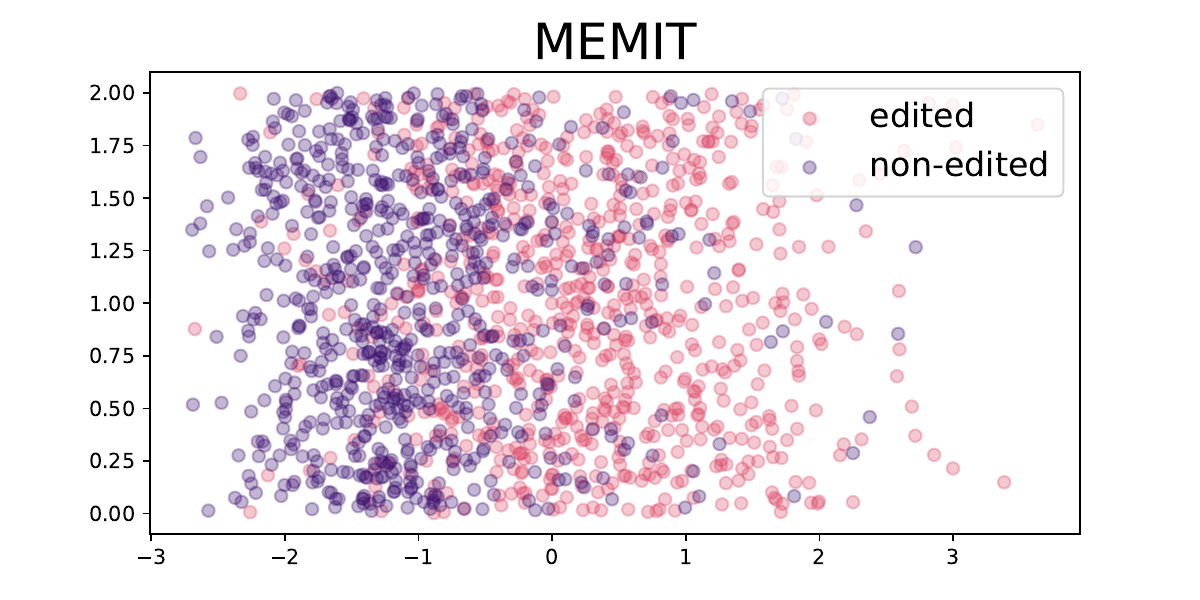}
\includegraphics[width=0.18\textwidth, trim={1cm 0cm 1cm 0.3cm}, clip]{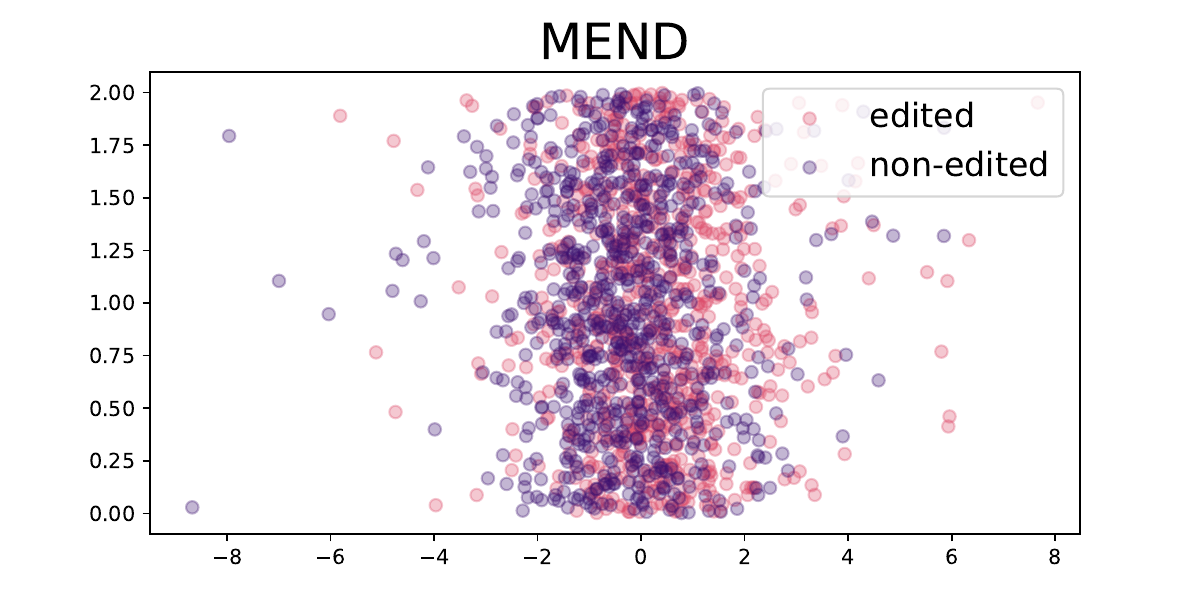}
\includegraphics[width=0.18\textwidth, trim={1cm 0cm 1cm 0.3cm}, clip]{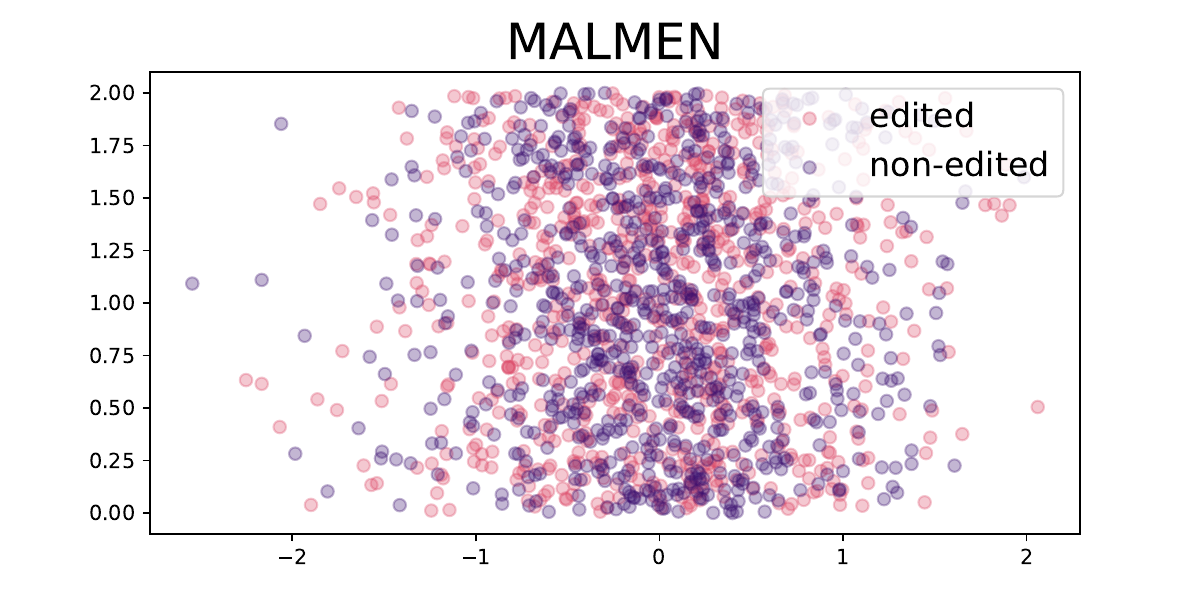}
\caption{Representation of edited and unedited facts (\zsre, GPT-J). X-axis: 1D LDA projection, Y-axis: random values to reduce point overlap.}
\label{fig:lda}
\end{figure*}

\paragraph{HS visualization.}To see how editing affects the hidden state representations,  we use Linear Discriminant Analysis (LDA)~\cite{duda2000pattern} to visualize the hidden state representations. Specifically, for each data point (fact) from the test set, we project the hidden state representation of the last token of $p'(s,r)$ into a one-dimensional space. We use the representations from the last layer of the respective LLM. We assign the one-dimensional representations random values on the y-axis to reduce point overlap. As shown in Figure~\ref{fig:lda}, the edited and unedited representations are not distinguishable in the unedited model. Editing with ROME and to less extent with MEMIT makes the representations more separable. We attribute this separability to the high confidence of LLMs in the edited facts, as shown later in Section~\ref{sec:baseline_evaluation}. Editing with MEND and MALMEN keeps the representations linearly inseparable.\footnote{We also experimented with PCA and t-SNE, but the representations were still inseparable.} This suggests that meta-learning KEs cause subtle changes on representations when editing facts. Similar patterns are observed on all datasets and methods (see Figure~\ref{fig:app_lda} in the Appendix).

\begin{figure*}[ht!]
\centering
  \includegraphics[width=.18\textwidth, trim={0.5cm 0cm 0.5cm 0.5cm}, clip]{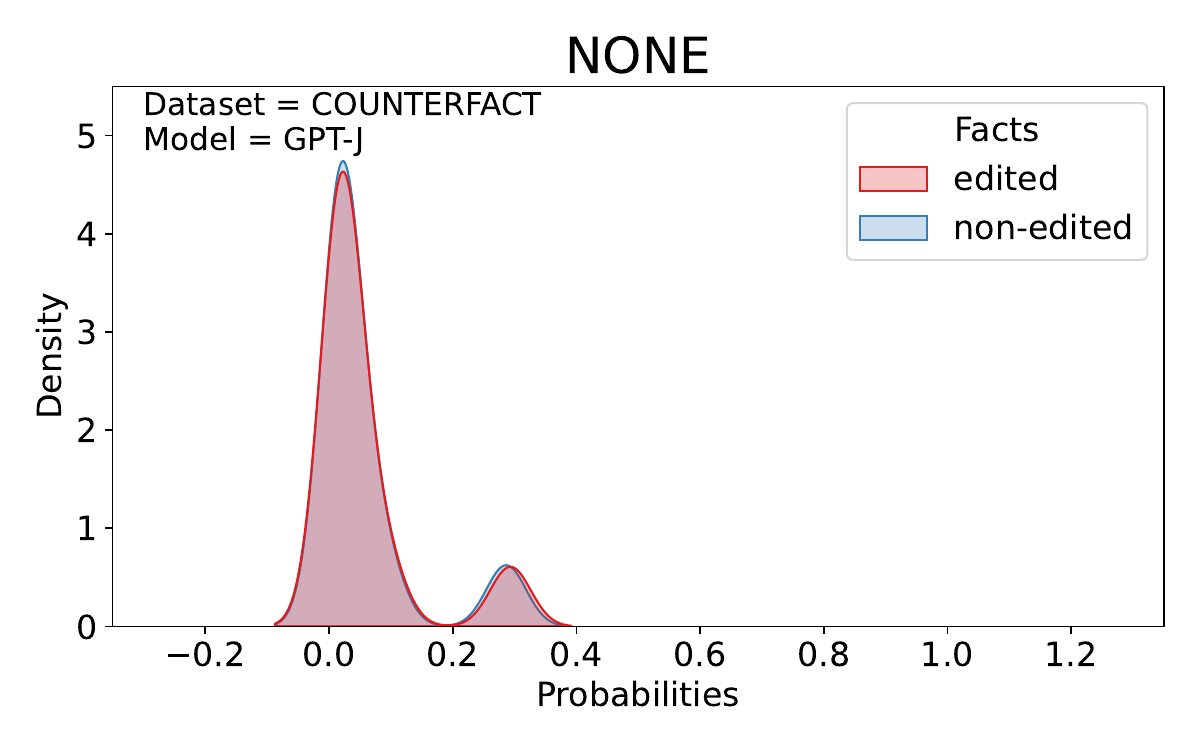}
  \includegraphics[width=.18\textwidth, trim={0.5cm 0cm 0.5cm 0.5cm}, clip]{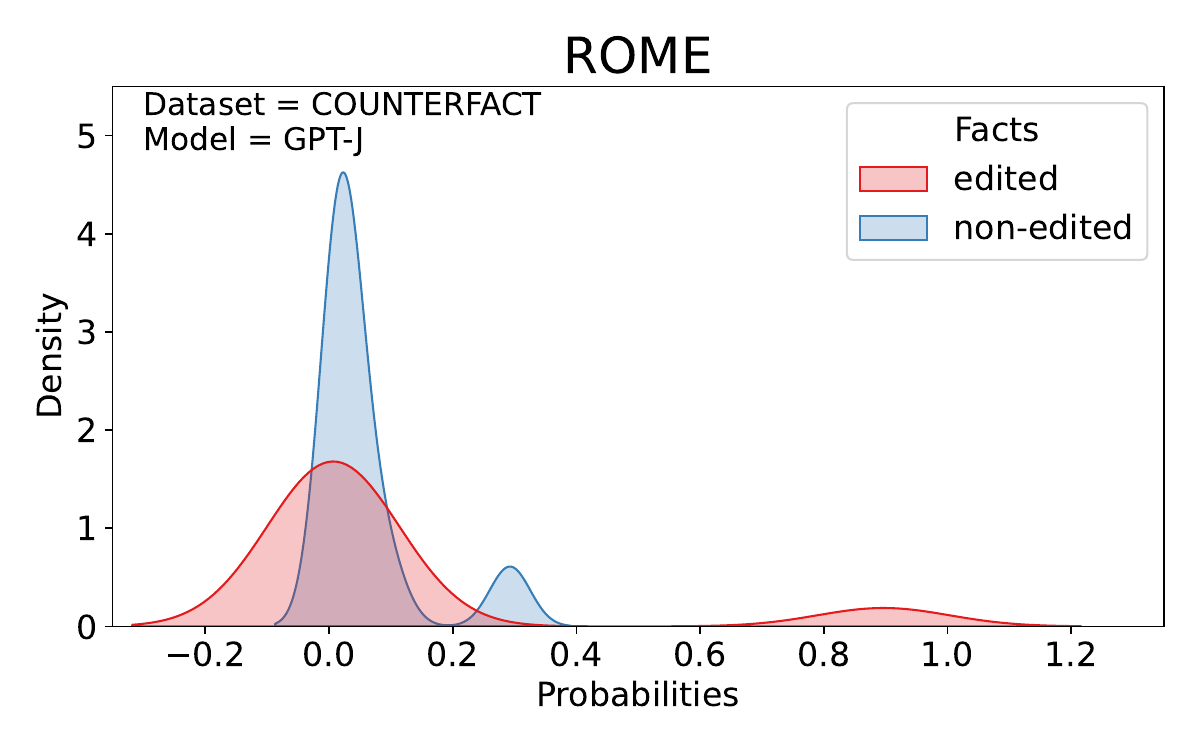}
  \includegraphics[width=.18\textwidth, trim={0.5cm 0cm 0.5cm 0.5cm}, clip]{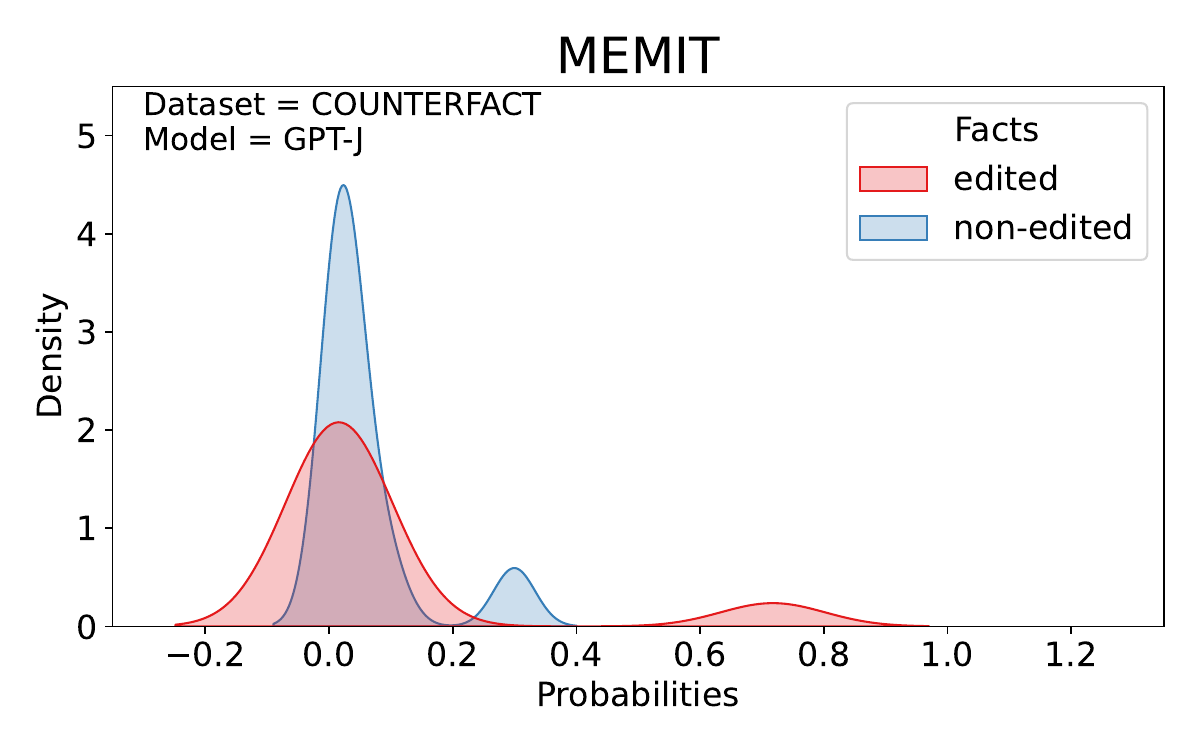}
  \includegraphics[width=.18\textwidth, trim={0.5cm 0cm 0.5cm 0.5cm}, clip]{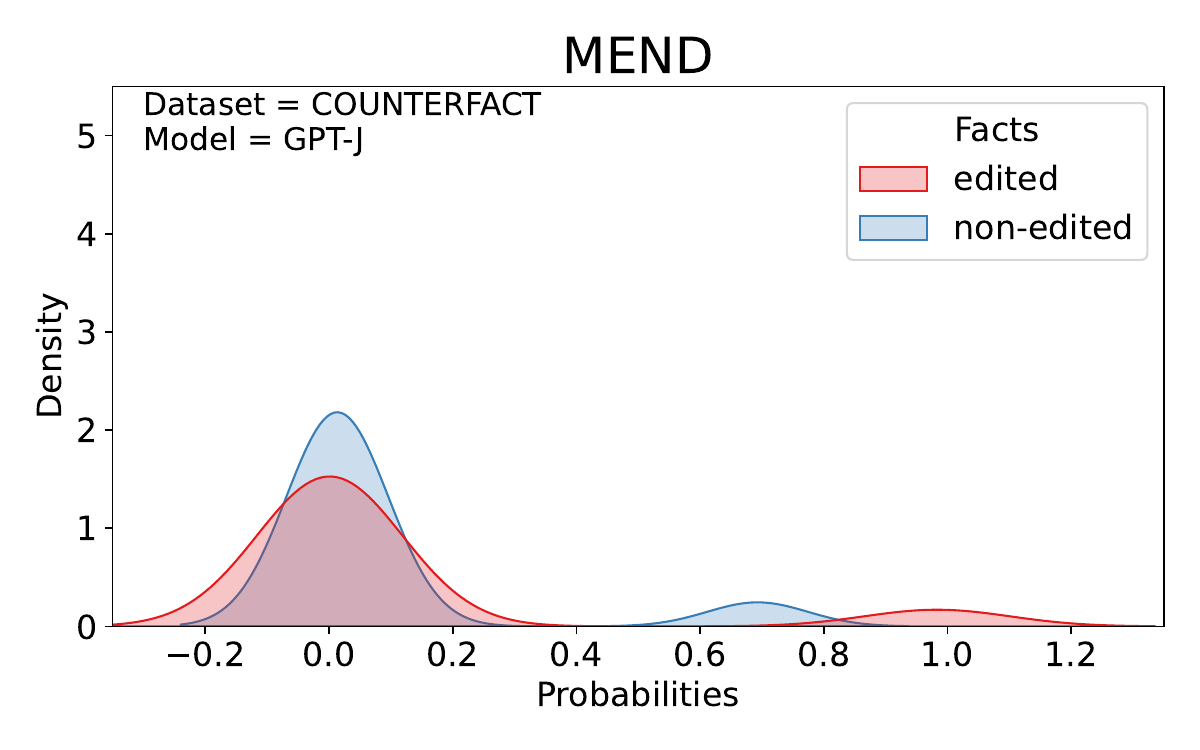}
  \includegraphics[width=.18\textwidth, trim={0.5cm 0cm 0.5cm 0.5cm}, clip]{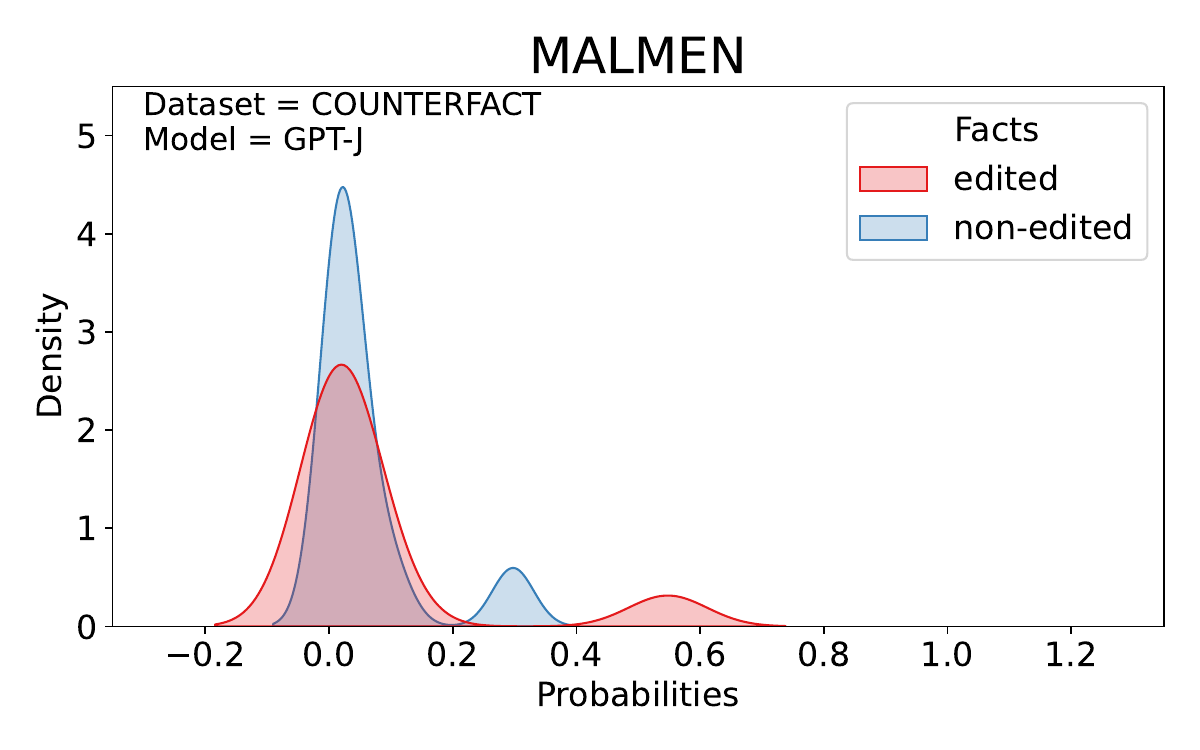}
\caption{Kernel density estimation for averaged top-10 probabilities for the next token (GPT-J, \cfd). }
\label{fig:kde}
\end{figure*}

\paragraph{PD visualization.} We investigate the output probability distributions for the next token from LLMs when prompting for edited and unedited facts with prompts that contain subjects and relations: $p'_1(s_1,r_1)...p'_n(s_n,r_n)$. We consider a balanced set consisting of the same number of edited and unedited facts. We analyze the probability distribution of each set by calculating the mean of the top 10 probabilities and plotting the kernel density estimate (KDE). We show the KDEs for GPT-J and \cfd in Figure~\ref{fig:kde}. The figure shows almost identical distributions of low probabilities for both edited and unedited facts in the unedited model NONE. We notice two peaks for both distributions in NONE: the left one is close to zero, while the right one is around 0.3. As we edit with different methods, we notice that the right peak for edited facts is shifted to the right, while the first one flattens, as the probabilities there become less frequent. The increase in probabilities varies depending on the method; ROME and MEND cause the highest increase, followed by MEMIT and MALMEN, which are used for mass-editing. We also notice that MEND increases the probabilities of unedited facts as well. Editing with MALMEN is the smoothest, since its probability distribution is the closest to that of the unedited facts. This potentially makes facts edited with MALMEN more difficult to detect.

\subsection{DEED Baselines}
\label{subsec:baselines}
Given a fact $(s, r, o, p')$, we prompt the given LLM $\mathcal{M}$, which consists of $L$ transformer blocks, with the prompt $p'(s,r)$ that consists of $N$ tokens $\{u_1,\ldots,u_N\}$. We use the hidden state representations \hs from the last transformer block and the last input token $h_{N,L} = \mathcal{M} [ p'(s,r) ]$ as a first feature for detecting edited facts. 
 We use the information from the last token, since this is the position where the information about the subject and the relation is condensed~\cite{geva-etal-2023-dissecting}.
As a second feature, we use \pd, the top-1000 probabilities of the next token $q = (q_1, \ldots, q_{1000})$ from $\mathcal{M}$ on the same prompt $p'(s,r)$.

\subsection{Evaluation}\label{sec:baseline_evaluation}

\begin{figure}[h]
\centering
  \includegraphics[scale=0.09, trim={0cm 0cm 0cm 3cm}, clip]{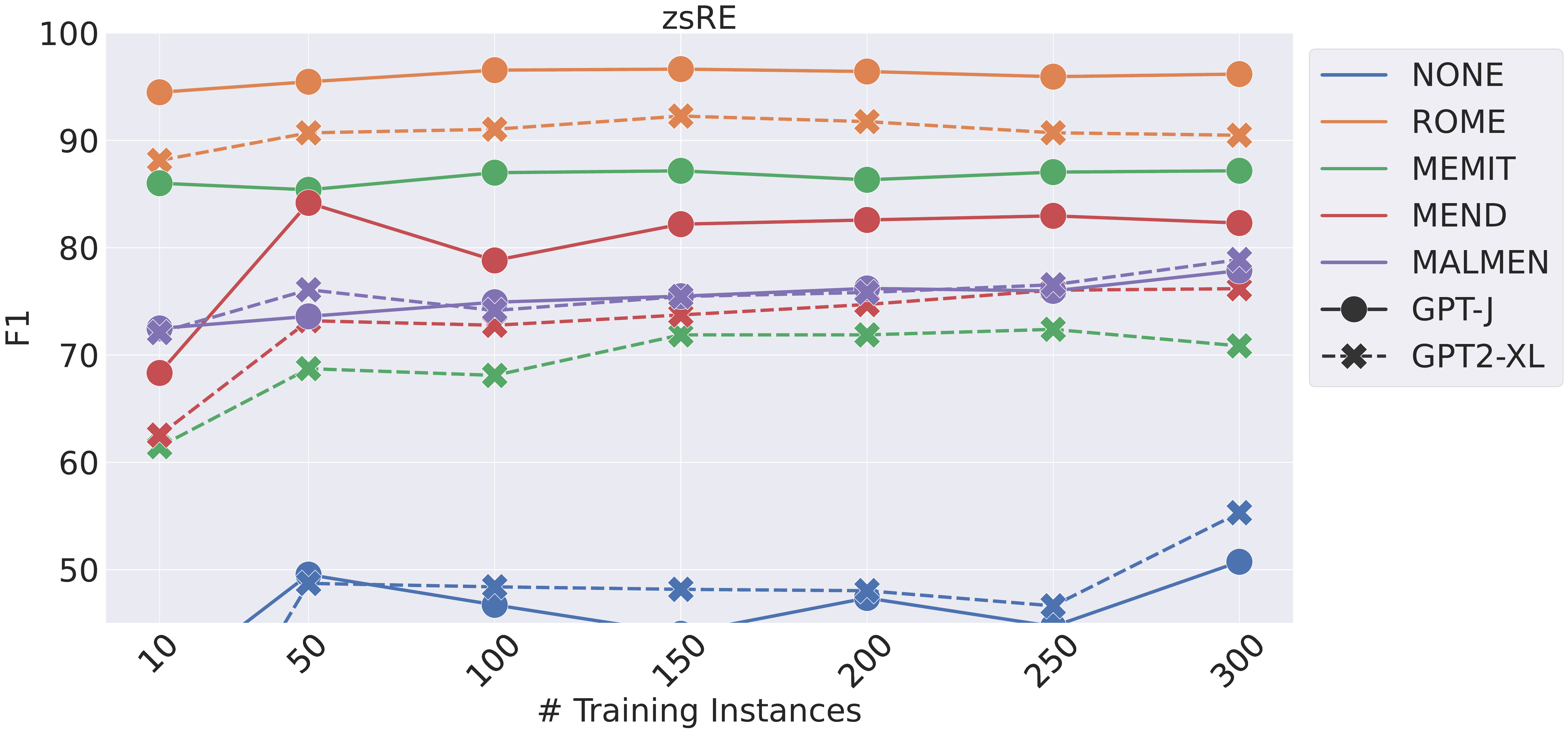}%
\caption{Detection performance with \pd on \zsre when varying the training set size.}
\vspace{-.3cm}
\label{fig:classification_performance_over_data:zsre}
\end{figure}

We train an AdaBoost classifier~\citep{freund1997decision} with a decision tree as a base classifier using both features, \hs and  \pd, on the DEED datasets (Section~\ref{subsec:deed_dataset}). 
We show the classification performance of training on 126 instances in the first part of Table~\ref{tab:classification_performance}. The minimum number of edits under all settings was 127 (cf.~Table~\ref{tab:edits_stats}). We use half of the edited facts (63) for training, and the rest for testing. Balancing the edited 63 facts with 63 unedited facts yields a training set of 126 examples. We keep the size of the training set constant across all settings for fair comparison, and experiment later with smaller and larger training sets (cf. Figure~\ref{fig:classification_performance_over_data:zsre}). As a sanity check, we train a classifier on the same sets of edited and unedited facts using features from the base \emph{unedited} model. We refer to this classifier as NONE. The classifier NONE shows a fair coin performance on the three metrics, ranging from 46.2 (GPT-J, \cfd, F1) to 49.8 (GPT2-XL, \zsre, F1). This verifies the correctness of our setup. We report precision, recall, and F1.

\paragraph{Locate-and-edit KEs vs. meta-learning KEs.} We find that the classifier is able to detect edits by locate-and-edit KEs (LE-KEs), i.e., ROME and MEMIT, relatively well, achieving a minimum F1 score of 74.2\% (MEMIT, GPT2-XL, \zsre). The highest F1 score of 98\% is achieved on (ROME, GPT-J, \zsre). On meta-learning KEs (ML-KEs), i.e., MEND and MALMEN, we observe lower performance in general. On MEND the performance varies between 53.6\% (GPT2-XL, \cfd) and 81.4\% (GPT-J, \zsre), whereas on MALMEN the gap between the lowest and highest performance is smaller than the same gap on MEND (62\% (GPT2-XL, \cfd), and  76\% (GPT-J, \zsre)). In general, this shows that \emph{edits, from the more computationally efficient LE-KEs, are easier to detect}. One reason for this discrepancy in performance between these two classes of methods might be that ML-KEs induce changes that generally target the last few layers, while LE-KEs change middle layers. Note for example that MEND increases the probabilities for both edited and unedited facts, while MALMEN slightly increases the probabilities for edited facts (cf. Figure~\ref{fig:kde}). This keeps the distance between edited and unedited facts minimal under ML-KEs, while LE-KEs conduct surgical changes that increase the probability for edited facts only. This can also be confirmed by the fact that the effects of LE-KEs on unrelated facts are less pronounced compared to ML-KEs (cf. Table~\ref{tab:edit_score}, Appendix).  %

\textbf{Ablation.} We conduct an ablation study, where we train only on \hs, or \pd (second and third part of Table~\ref{tab:classification_performance} respectively). We notice that using \hs leads to low performance on the ML-KEs methods, where the F1 score varies between 48.5\% (MALMEN, GPT2-XL, \zsre) and 62.6\% (MALMEN, GPT-2XL, \cfd). On LE-KEs, \hs still performs well with a minimum F1 score of 65.3\% (MEMIT, GPT-J, \cfd). In contrast, \emph{detecting edits with \pd is more robust and performs well on LE-KEs and ML-KEs}. The minimum F1 scores with \pd is 65.2\% (MALMEN, GPT-J, \cfd). \pd's robustness might be explained by the fact that it is derived from a late position in the LLM, namely the language modeling head, which is responsible for the final outputs. Another important observation is that using \pd individually can lead to better performance than combining it with \hs in some cases (e.g., the F1 score on (MEND, GPT2-XL, \cfd) is 53.6\% when using both features, but 73\% when using only \pd). This observation suggests that \pd and \hs could be combined in better ways. We leave exploring the interplay between \hs and \pd to future work. We further explore using \hs from different layers in Appendix~\ref{app:hs_layers}, observing that the representations start becoming separable after the edited layers, and that the representations from some layers (e.g., the penultimate layer) perform better than the representations of the last layer.

\begin{table*}[h!]
\centering
\resizebox{\textwidth}{!}{%
\begin{tabular}{ll@{\hskip 5mm}lll@{\hskip 7mm}lll@{\hskip 10mm}lll@{\hskip 7mm}lll@{\hskip 10mm}lll@{\hskip 7mm}lll}

\toprule

&  
& \multicolumn{6}{l}{\textbf{Hidden States + Prob. Dist. (HS + PD)}} 

& \multicolumn{6}{c}{\textbf{Hidden States (HS)}} 
& \multicolumn{6}{c}{\textbf{Probability Distribution (PD)}}\\ 
& 
& \multicolumn{3}{c}{\zsre} & \multicolumn{3}{l}{\cfd} 
& \multicolumn{3}{c}{\zsre} & \multicolumn{3}{l}{\cfd} 
& \multicolumn{3}{c}{\zsre} & \multicolumn{3}{l}{\cfd}\\ \midrule
 & KE &  Pr. &  Rec. &  F1   &  Pr. &  Rec. &  F1 &  Pr. &  Rec. &  F1   &  Pr. &  Rec. &  F1 &  Pr. &  Rec. &  F1   &  Pr. &  Rec. &  F1 \\ \midrule
\multirow{5}{*}{\rotatebox[origin=c]{90}{GPT-J}} 
&   NONE & 49.4 &       46.6 &   48.0 &          48.6 &       44.1 &   46.2 &          49.4 &       46.6 &   48.0 &          48.6 &       44.1 &   46.2 &          50.7 &       42.6 &   46.3 &          51.4 &       49.6 &   50.5 \\
&   ROME & 98.6 &       97.4 &   \textbf{98.0} &          91.2 &       90.2 &   \underline{90.7} &          98.0 &       95.6 &   96.8 &          87.1 &       84.9 &   86.0 &          98.5 &       94.5 &   96.4 &          89.3 &       89.2 &   89.2 \\
&  MEMIT & 87.7 &       89.8 &   \textbf{88.7} &          79.0 &       76.7 &   \underline{77.9} &          79.0 &       81.6 &   80.2 &          70.8 &       60.6 &   65.3 &          83.8 &       89.8 &   86.7 &          78.0 &       73.4 &   75.6 \\
&   MEND & 78.4 &       84.6 &   81.4 &          59.9 &       58.8 &   59.4 &          61.6 &       59.3 &   60.4 &          59.9 &       58.8 &   59.4 &          78.3 &       85.9 &   \textbf{81.9} &          63.5 &       96.1 &   \underline{76.5} \\
& MALMEN & 76.6 &       75.3 &   \textbf{76.0} &          62.0 &       62.0 &   62.0 &          49.9 &       49.6 &   49.7 &          57.3 &       56.8 &   57.1 &          71.2 &       77.1 &   74.0 &          66.8 &       63.7 &   \underline{65.2} \\
 \midrule
\multirow{5}{*}{\rotatebox[origin=c]{90}{GPT2-XL}}
&   NONE & 49.7 &       50.0 &   49.8 &          51.4 &       48.0 &   49.6 &          49.3 &       44.8 &   47.0 &          51.4 &       48.0 &   49.6 &          51.0 &       50.2 &   50.6 &          49.6 &       48.0 &   48.8 \\
&   ROME & 95.0 &       98.2 &   \textbf{96.6} &          91.8 &       93.6 &   \underline{92.7} &          94.7 &       97.6 &   96.1 &          91.3 &       92.1 &   91.7 &          94.6 &       90.2 &   92.3 &          87.6 &       91.4 &   89.4 \\
&  MEMIT & 77.7 &       70.9 &   \textbf{74.2} &          76.0 &       73.2 &   74.5 &          71.8 &       68.2 &   69.9 &          69.4 &       66.7 &   68.0 &          69.5 &       65.5 &   67.5 &          77.3 &       75.1 &   \underline{76.2} \\
&   MEND & 67.0 &       78.9 &   72.5 &          55.1 &       52.2 &   53.6 &          62.6 &       53.7 &   57.8 &          54.3 &       50.8 &   52.5 &          66.4 &       88.0 &   \textbf{75.7} &          60.7 &       91.5 &   \underline{73.0} \\
& MALMEN & 69.8 &       71.0 &   \textbf{70.4} &          68.2 &       67.2 &   \underline{67.7} &          49.3 &       47.7 &   48.5 &          61.2 &       64.1 &   62.6 &          71.2 &       65.0 &   68.0 &          64.2 &       67.2 &   65.6 \\
 \bottomrule
\end{tabular}%
}
\caption{Detection performance on \zsre and \cfd with different features sets. The training set has 126 instances in all settings. The best F1 across all feature sets is in \textbf{bold} for \zsre, and \underline{underlined} for \cfd. The number of test instances is shown in Table~\ref{tab:data_stats} in the Appendix.}
\label{tab:classification_performance}
\end{table*}

\paragraph{Training set size.} 
Figure \ref{fig:classification_performance_over_data:zsre} shows the performance of different training set sizes with \pd. Exhaustive results are presented in Table~\ref{tab:cls_data} in the Appendix. In general, the classification performance starts reaching a plateau after 50 training instances. For example, on MEMIT-edited facts in GPT-J (green circles), F1 reaches 85.4\% after 50 training instances, while the highest F1 of 87.2\% is achieved after 300 training data points. Therefore, we emphasize that \emph{this classifier remains effective even when the training data consists of less than one hundred examples. We also observe that facts that are edited with ML-KEs are not only harder to detect than facts edited with LE-KEs (as noted before), but also that scaling the size of the training set does not improve performance. }

\section{Cross-Domain DEED}
\label{sec:challenging}

\begin{figure*}[ht!]
    \centering
    \includegraphics[width=0.95\textwidth, trim= 5cm 0cm 5cm 0cm]{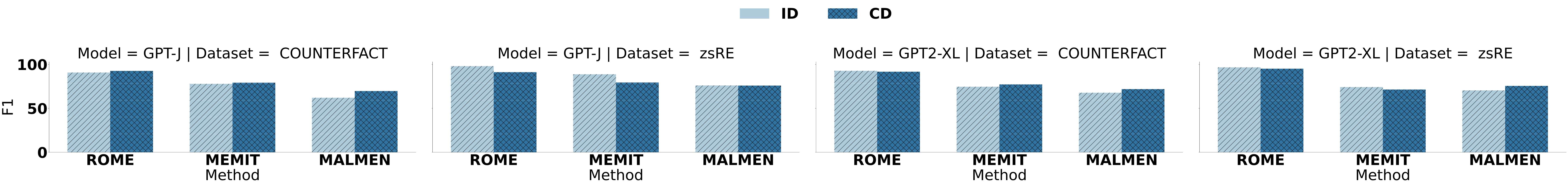}

    \caption{Comparison between in-domain (\indomain) and cross-domain (\crossdomain) classification performance. Classification is done based on \hs{} and \pd{}.}
    \label{fig:ood}
\end{figure*}

Obtaining a training set for the detection classifier requires knowing instances of edited facts in the edited model. However, this assumption may not always hold. Consider a scenario where an open-source model is further fine-tuned for a specific task of interest to the user, yet no explicit information is provided regarding the edited facts. Many models are fine-tuned with instructions or reinforcement learning from human feedback (RLHF), and are made publicly available~\cite{lai-etal-2023-okapi}. In these cases, we have access to the original model (from which we could obtain our training set), but do not know which facts have been edited into the fine-tuned model. To address such cases, we assess the performance of the classifier, which uses \hs{} and \pd{}, in a cross-domain setting, where the classifier is trained on representations from one edited model and then applied to detecting knowledge edits from a separate edited model (which is a fine-tuned version of the original model). The \crossdomain edited models have the same architecture, but different weights. To simulate this scenario, we use representations from edited GPT-J and GPT2-XL to train the classifier and test it in detecting knowledge edits from instruction-tuned versions of GPT-J\footnote{\url{https://huggingface.co/togethercomputer/GPT-JT-6B-v1}} and GPT2-XL.\footnote{\url{https://huggingface.co/lgaalves/gpt2-xl_lima}} 
Note that the facts in the training and test sets are still disjoint. For this experiment, we consider ROME, MEMIT and MALMEN.

As shown in Figure~\ref{fig:ood}, \emph{the performance on cross-domain edits (\crossdomain) is on par with performance on in-domain data (\indomain)}. For example, on \zsre, \crossdomain F1 on ROME-edited facts on GPT2-XL lima is 95\%, similar to the \indomain F1 of 96.6\%. Full results with \hs/\pd are shown in Table~\ref{tab:ood} in the Appendix. This makes our DEED baseline highly practical, as it can be leveraged to detect edits in fine-tuned models, while making use of training data that is derived from their counterpart original models.

\paragraph{Cross-dataset generalization.} The results of our previous experiment indicate that detection classifiers trained on edits in one model generalize to edits in other models of the same family. To further explore out-of-distribution scenarios, we impose an additional constraint by examining whether detection classifiers trained on edits from one dataset (e.g., \cfd{}) generalize to edits from another dataset (e.g., \zsre). Here, we maintain the assumption from our previous experiment that the base LLM is known, but introduce a new constraint: the dataset used to conduct edits on the test LLM is unknown, necessitating training on one dataset and testing on another. The primary objective of this experiment is to investigate whether edit detection classifiers generalize effectively across datasets.

As shown in Table~\ref{tab:cross_data} the detection classifiers generalize well to unseen datasets in most cases. One exception is generalization from \zsre to \cfd with MALMEN, where we see a dramatic decrease in performance. One reason for this might be the smooth changes induced by MALMEN (cf. Section~\ref{subsec:analysis}). These changes are dataset-specific, which makes generalization more difficult (cf. Figure~\ref{fig:kde:cfd:gpt2} and~\ref{fig:kde:zsre:gpt2} in the Appendix).

\begin{table}[ht!]

\centering
\small
    \centering
    \begin{tabular}{lp{2cm}cc}
        \toprule
        & Method &      \zsre  &    \cfd \\
        \midrule
        \multirow{3}{*}{\rotatebox[origin=c]{90}{GPT-J}} 
        
        &ROME &   82.7 & 89.6 \\
        &MEMIT &   73.4 & 76.8 \\
        &MALMEN &  63.7 & 25.9 \\ \midrule
        \multirow{3}{*}{\rotatebox[origin=c]{90}{GPT2-XL}} 
        
         & ROME &    84.5 & 90.8 \\
         & MEMIT &    70.4 & 72.7 \\
        & MALMEN &    62.1 &  6.8 \\
            \addlinespace[0.17cm]
        \bottomrule

\end{tabular}
\caption{F1 scores in the cross-datasets setting. Classifiers are trained on \hs and \pd from one dataset and tested on another dataset in \crossdomain setting. Tests sets are shown in the header.}
\label{tab:cross_data}

\end{table}

\section{Conclusion}
\label{sec:conclusion}
In this paper, we introduced a novel task: detecting knowledge edits in language models. 
We see promising applications in uncovering potential malicious model modification and enhancing the transparency of LLMs' generations. 
We instantiated the task with several baselines, which use hidden state representations and probability distributions as input features. These baselines demonstrate a straightforward yet effective approach to the task. 
Our evaluation across two LLMs, four editing techniques (2 LE-KEs and 2 ML-KEs), and two datasets, shows that the classifiers can reliably detect edits, even in scenarios characterized by limited training data.  Moreover, we showed that these baselines generalize to cross-domain settings, where training and test data come from different models and datasets, highlighting their practical utility. %
Our work gives rise to several future work directions: 1) detecting if an LLM has undergone any edits at all; 2) studying detection generalization across KEs; 3) investigating more constrained settings, where one has only access to the LLM's outputs; 4) combating malicious edits.

\section{Limitations}
\label{sec:limitations}
\paragraph{Access to \hs and \pd.} The baseline detection classifier we proposed requires access to the hidden states or probability distribution. When using closed source LLMs, these features might not be available. Inspired by \citet{rosati-etal-2024-long}, we plan to use the model behaviors shift on prompts with similar semantics but in different lengths to separate edited and unedited knowledge.

\paragraph{Unknown editing methods.} In our current setting, we assume we know the editing method. However, the editing method might not be known and an edited model can be modified by multiple editing methods. In future work, we will investigate the generalization of our baseline classifiers between different KEs, and detecting edits in models that have edited with several KEs.

\paragraph{Number of edits.} Previous work shows that a larger number of edits often leads to a worse editing performance~\cite{meng-etal-2022-memit}. Although we only keep successful edits for the detection task, it is unknown whether and how the number of edits will impact the detection performance. We plan to study this aspect in future work.

\section*{Acknowledgment}
This research was partially funded by the German Federal Ministry of Education and Research (BMBF) as German Academic Exchange Service (DAAD) project DEED under Grant No. 30001797. This work was conducted while Zhixue Zhao was an AInet Fellow, funded under the DAAD Fellowship Program (No. 57733361).

\bibliography{anthology,tacl2021}
\bibliographystyle{acl_natbib}

\appendix

\appendix

\section{Complementary Results}

\begin{figure*}[h!]
    \begin{subfigure}[t]{0.32\textwidth}
        \centering

  \includegraphics[width=\columnwidth, trim={1cm 0cm 1cm 0.3cm}, clip]{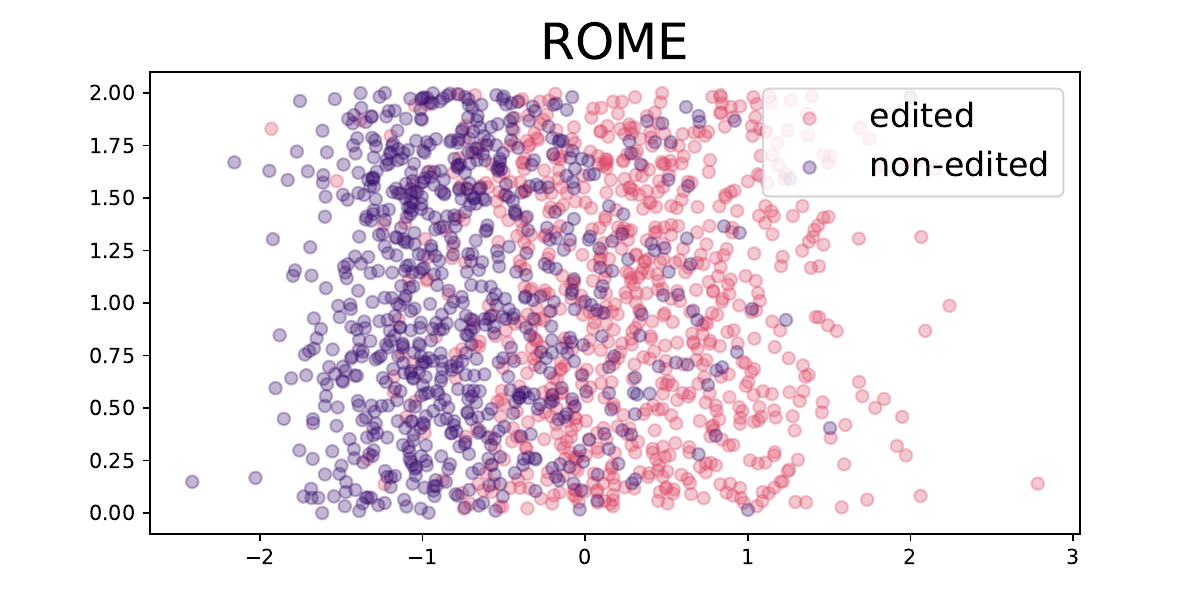}\\
  \includegraphics[width=\columnwidth, trim={1cm 0cm 1cm 0.3cm}, clip]{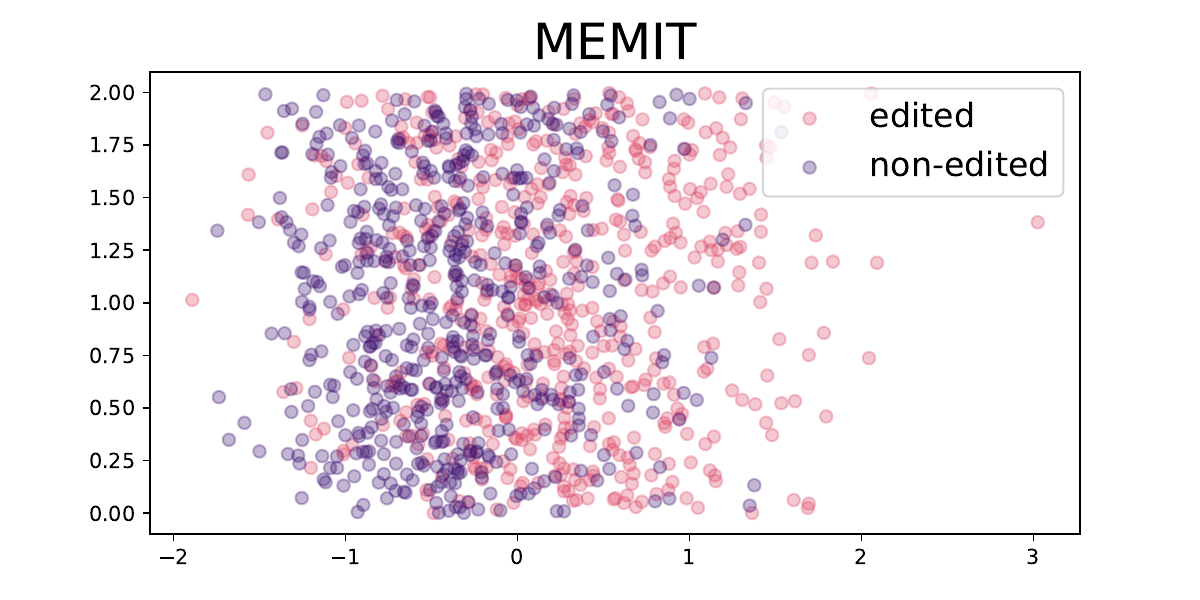}\\
  \includegraphics[width=\columnwidth, trim={1cm 0cm 1cm 0.3cm}, clip]{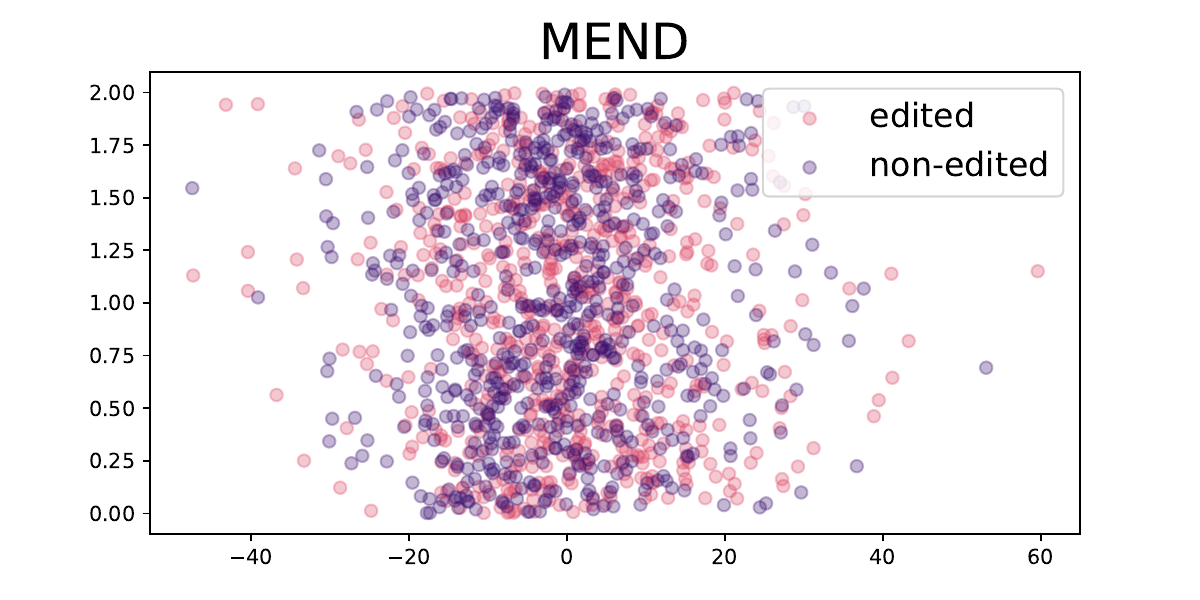}\\
  \includegraphics[width=\columnwidth, trim={1cm 0cm 1cm 0.3cm}, clip]{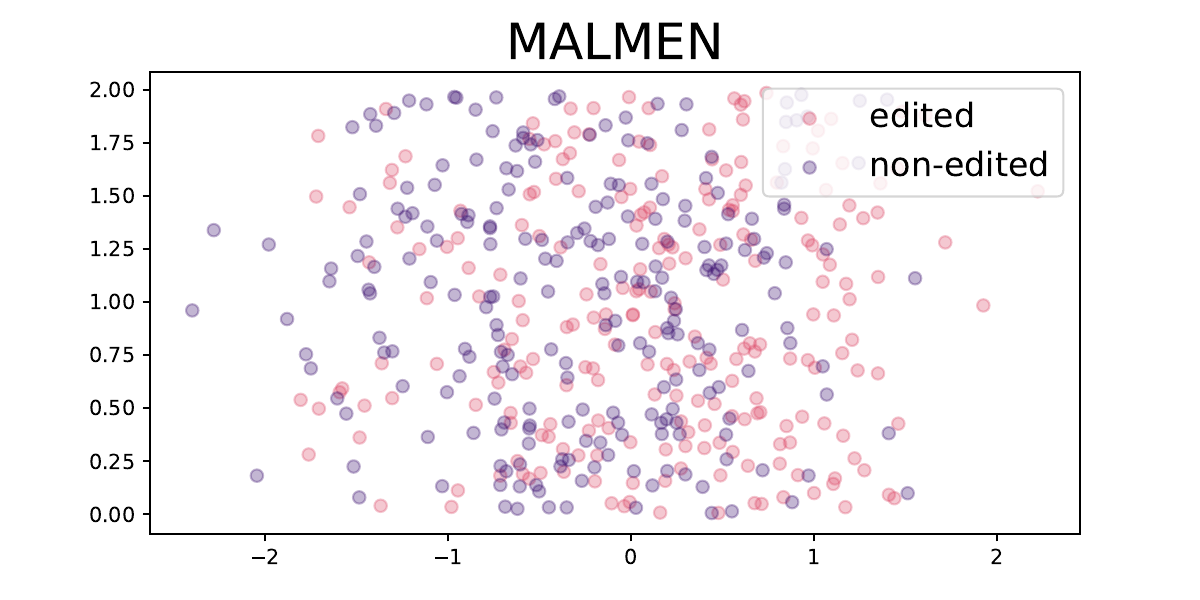}\\

\caption{\cfd and GPT-J.}
\label{fig:lda:cfd:gptj}
    \end{subfigure}
    \hfill
    \begin{subfigure}[t]{0.32\textwidth}
        \centering
  \includegraphics[width=\columnwidth, trim={1cm 0cm 1cm 0.3cm}, clip]{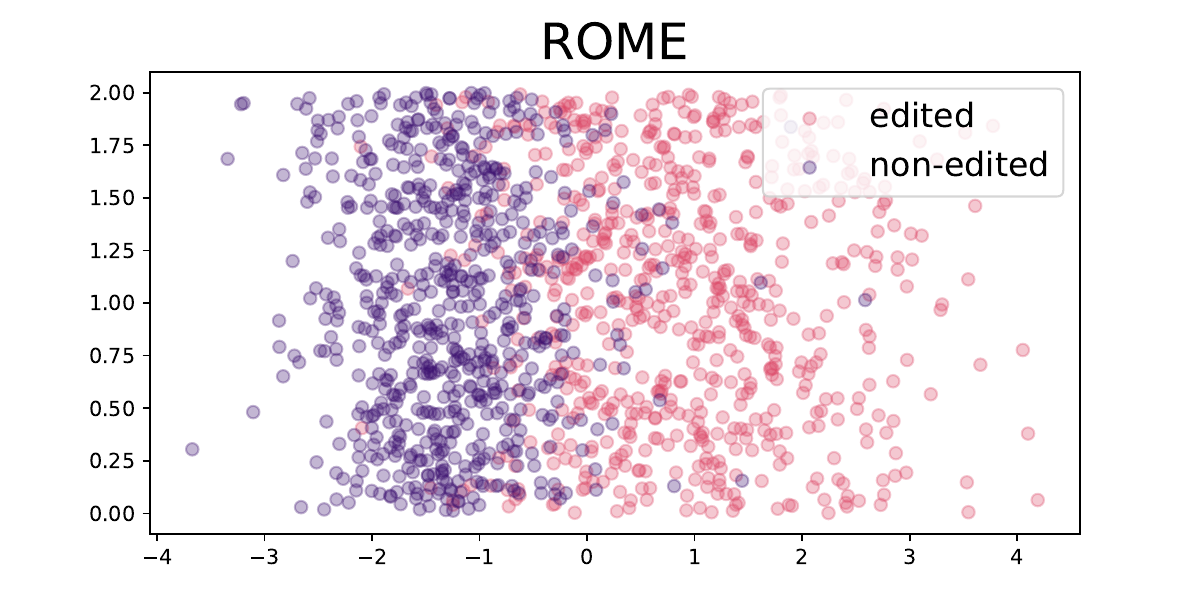}\\
  \includegraphics[width=\columnwidth, trim={1cm 0cm 1cm 0.3cm}, clip]{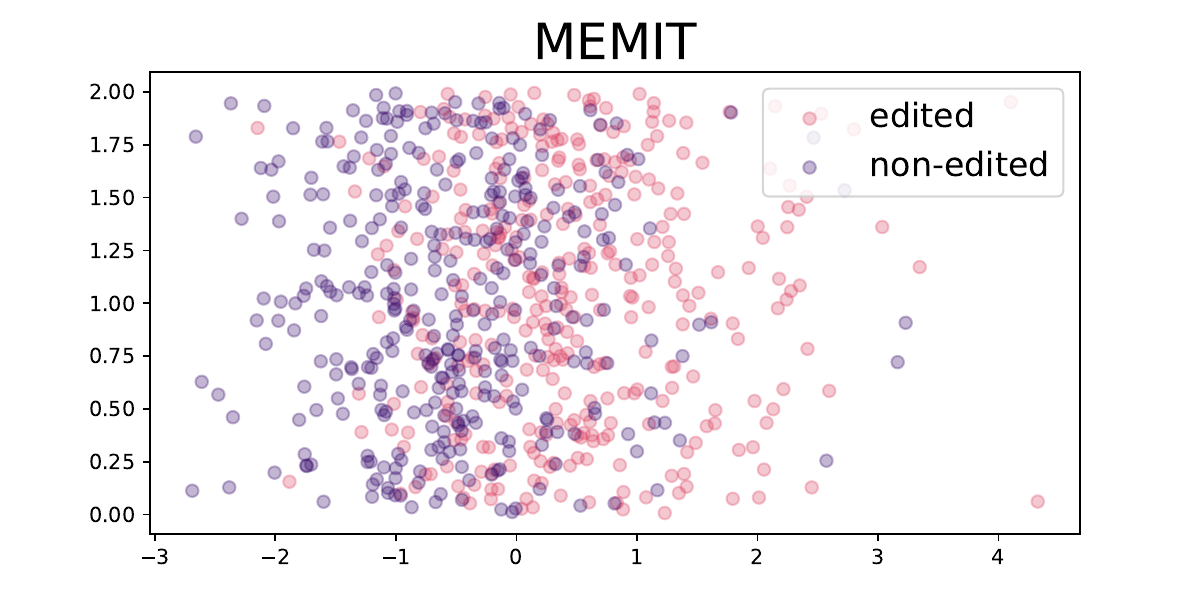}\\
  \includegraphics[width=\columnwidth, trim={1cm 0cm 1cm 0.3cm}, clip]{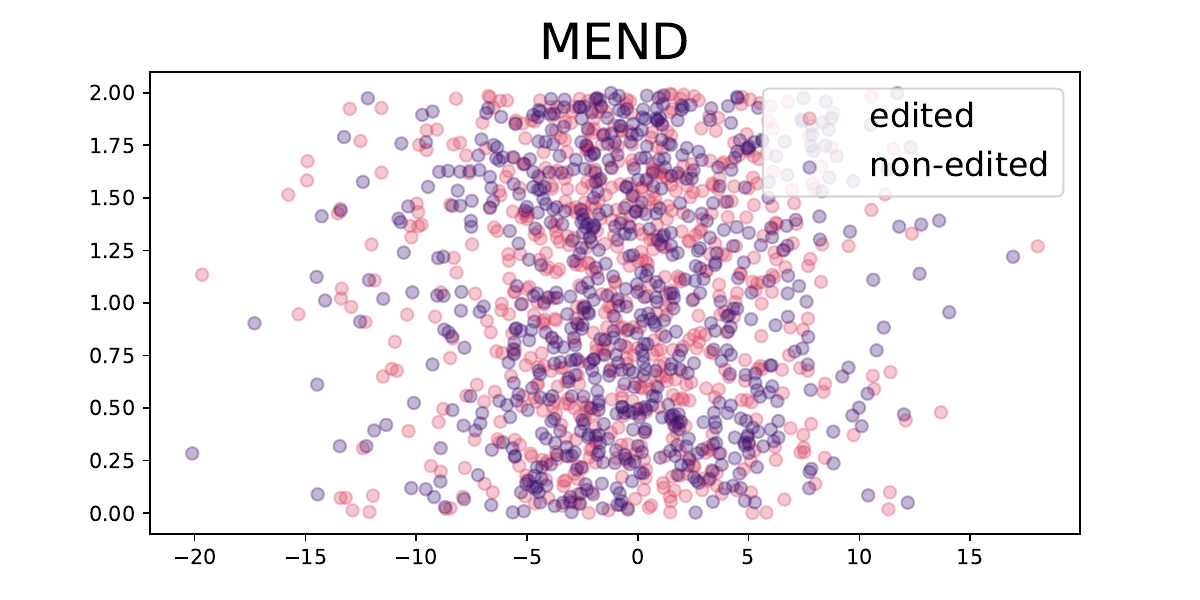}\\
  \includegraphics[width=\columnwidth, trim={1cm 0cm 1cm 0.3cm}, clip]{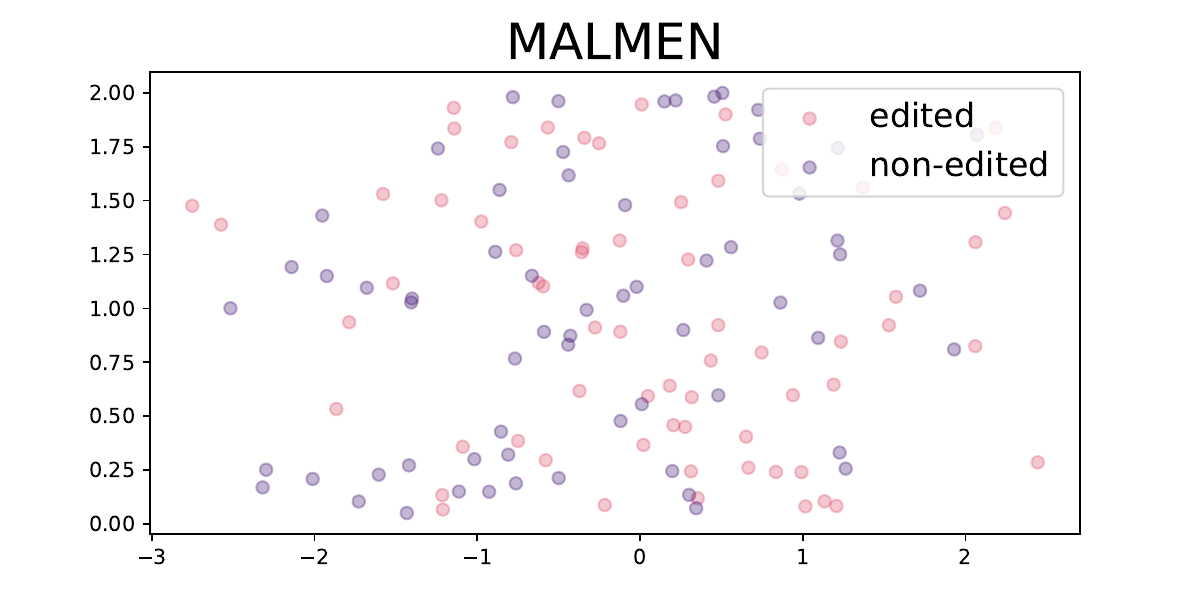}\\

\caption{\cfd and GPT2-XL.}
\label{fig:lda:cfd:gpt2}
    \end{subfigure}
    \hfill
    \begin{subfigure}[t]{0.32\textwidth}
        \centering
  \includegraphics[width=\columnwidth, trim={1cm 0cm 1cm 0.3cm}, clip]{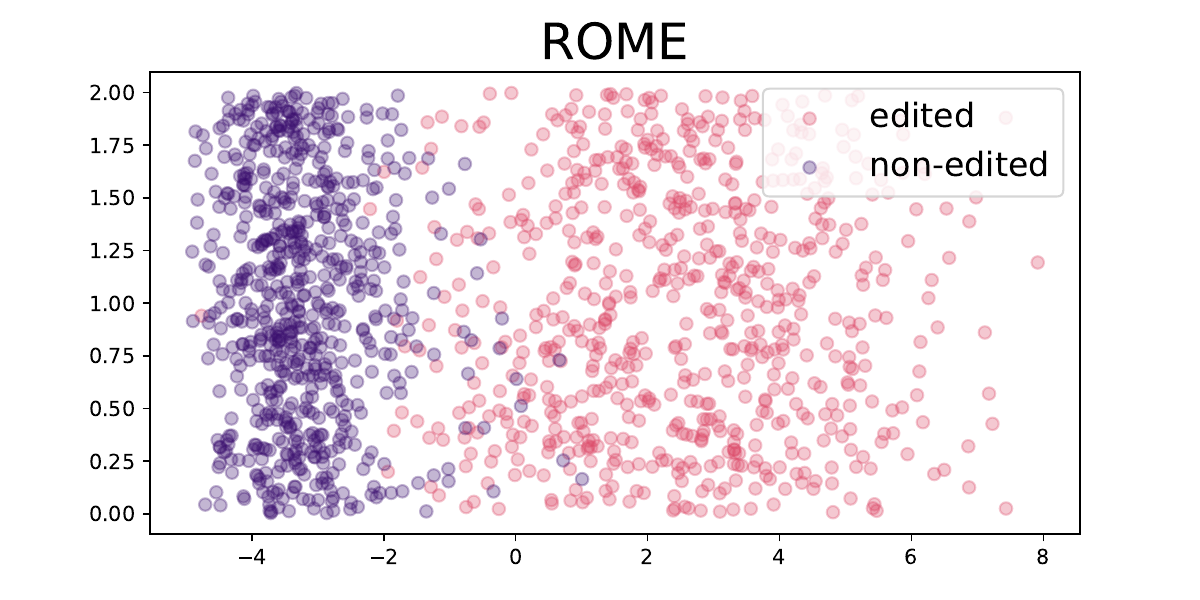}\\
  \includegraphics[width=\columnwidth, trim={1cm 0cm 1cm 0.3cm}, clip]{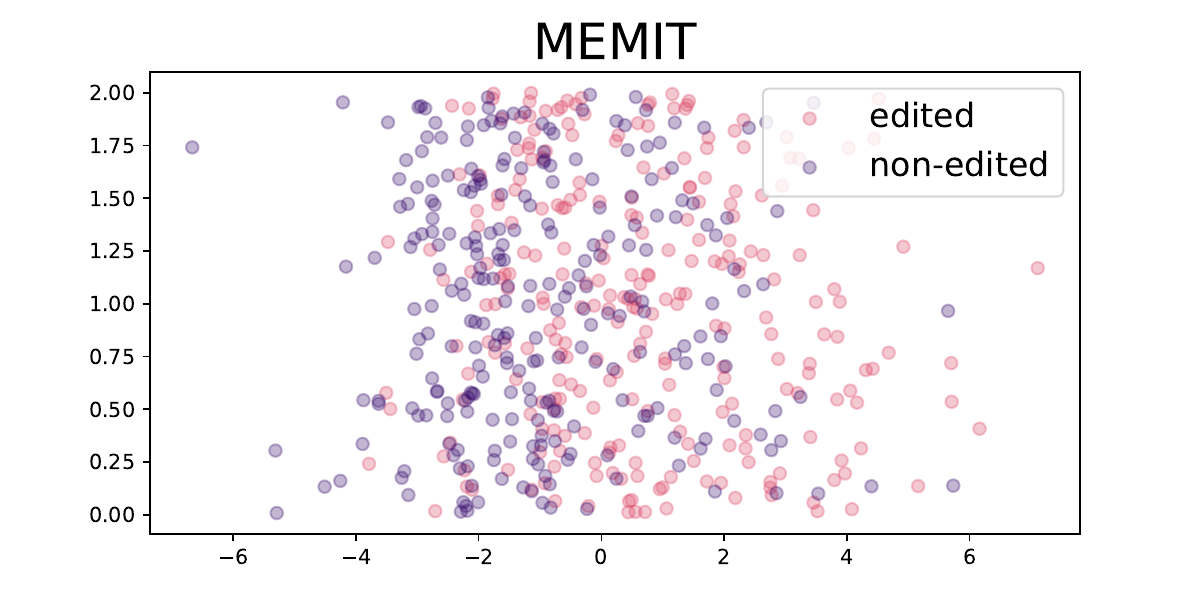}\\
  \includegraphics[width=\columnwidth, trim={1cm 0cm 1cm 0.3cm}, clip]{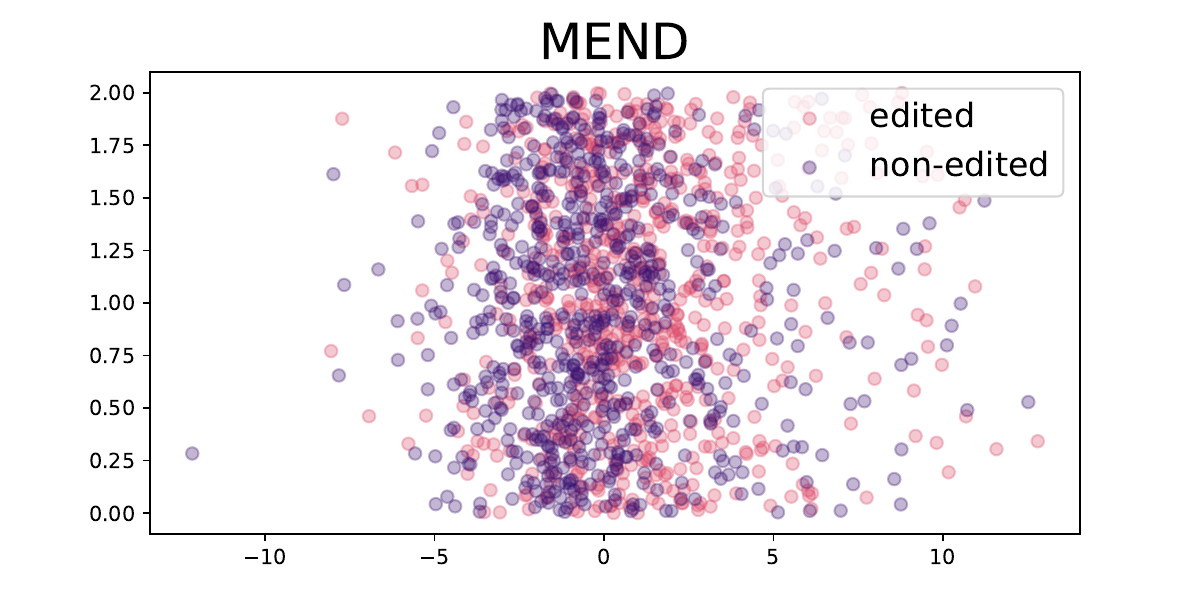}\\
  \includegraphics[width=\columnwidth, trim={1cm 0cm 1cm 0.3cm}, clip]{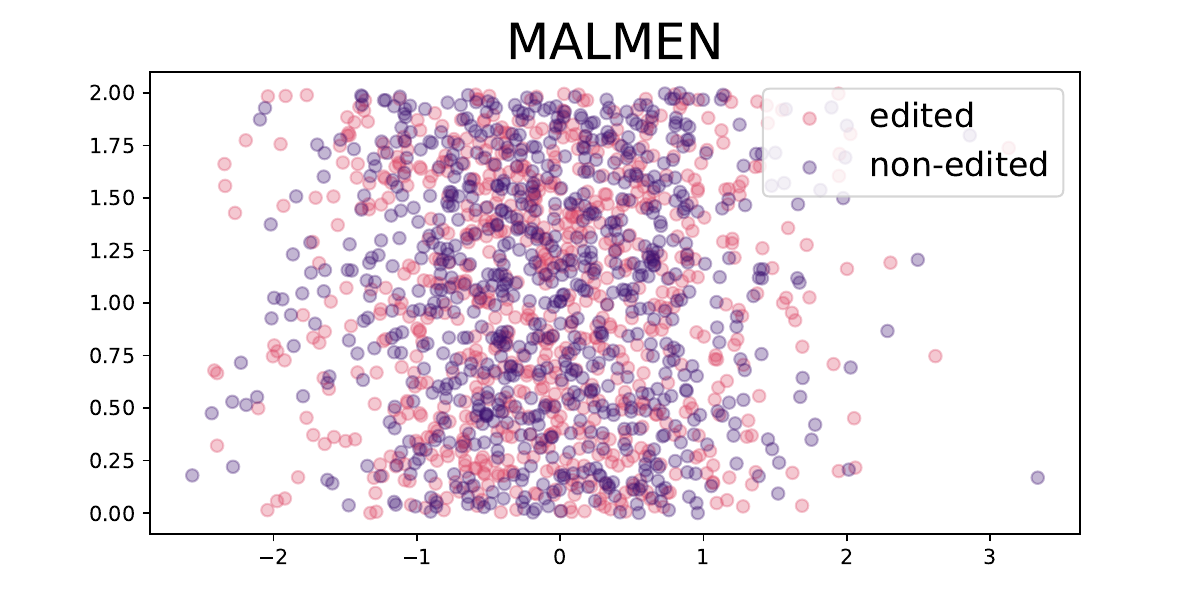}\\

\caption{\zsre and GPT2-XL.}
\label{fig:lda:zsre:gpt2}
    \end{subfigure}
    \caption{Respresentations of edited and non-edited facts.}
    \label{fig:app_lda}
\end{figure*}

\begin{figure}[h]
\centering

\includegraphics[scale=0.09, trim={0cm 0cm 0cm 3cm}, clip]{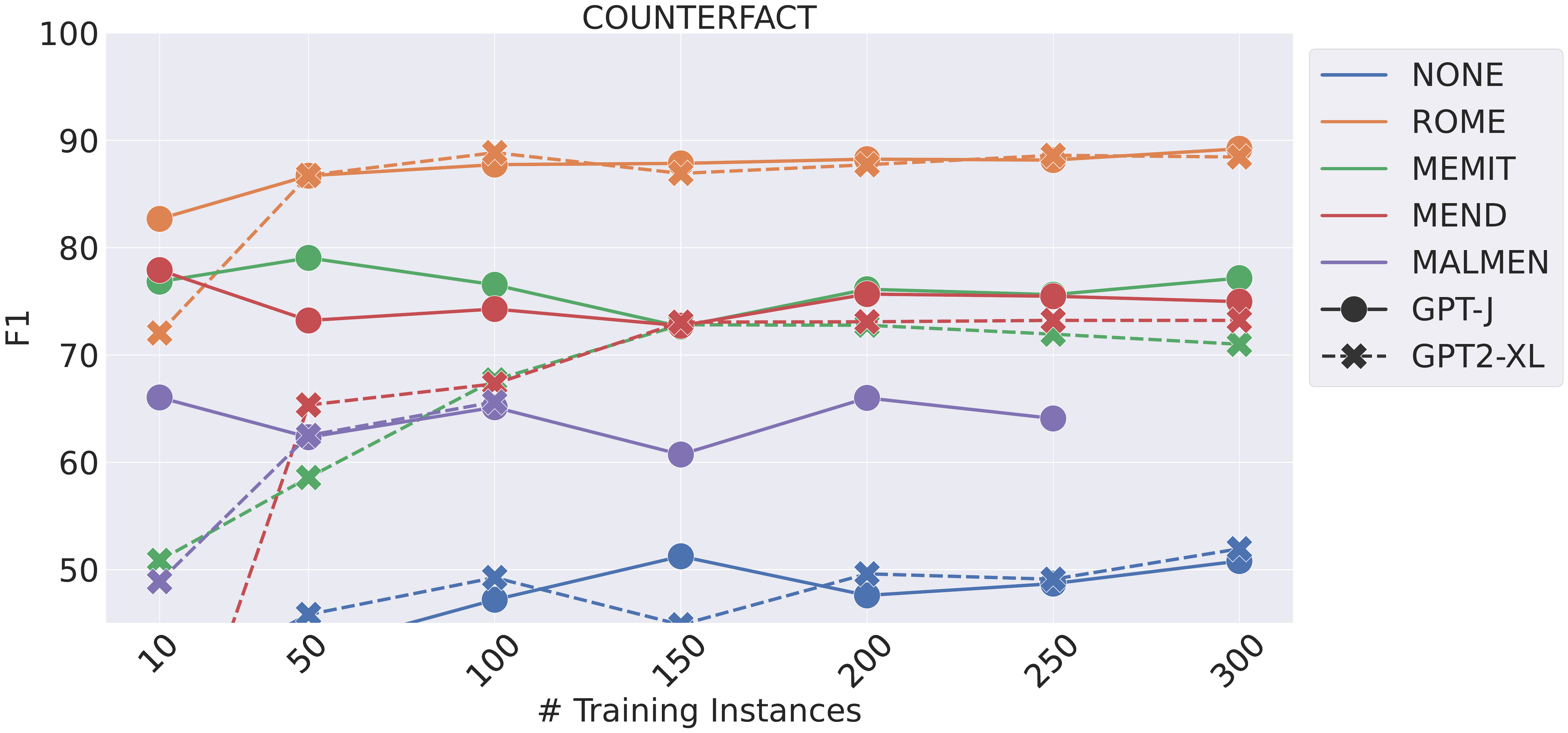}

\caption{Detection performance with \pd on \cfd when varying training set size.}
\label{fig:classification_performance_over_data:cfd}
\end{figure}

\begin{figure*}[h!]
    \begin{subfigure}[t]{0.32\textwidth}
        \centering

  \includegraphics[width=\columnwidth, trim={0.5cm 0cm 0.5cm 0.3cm}, clip]{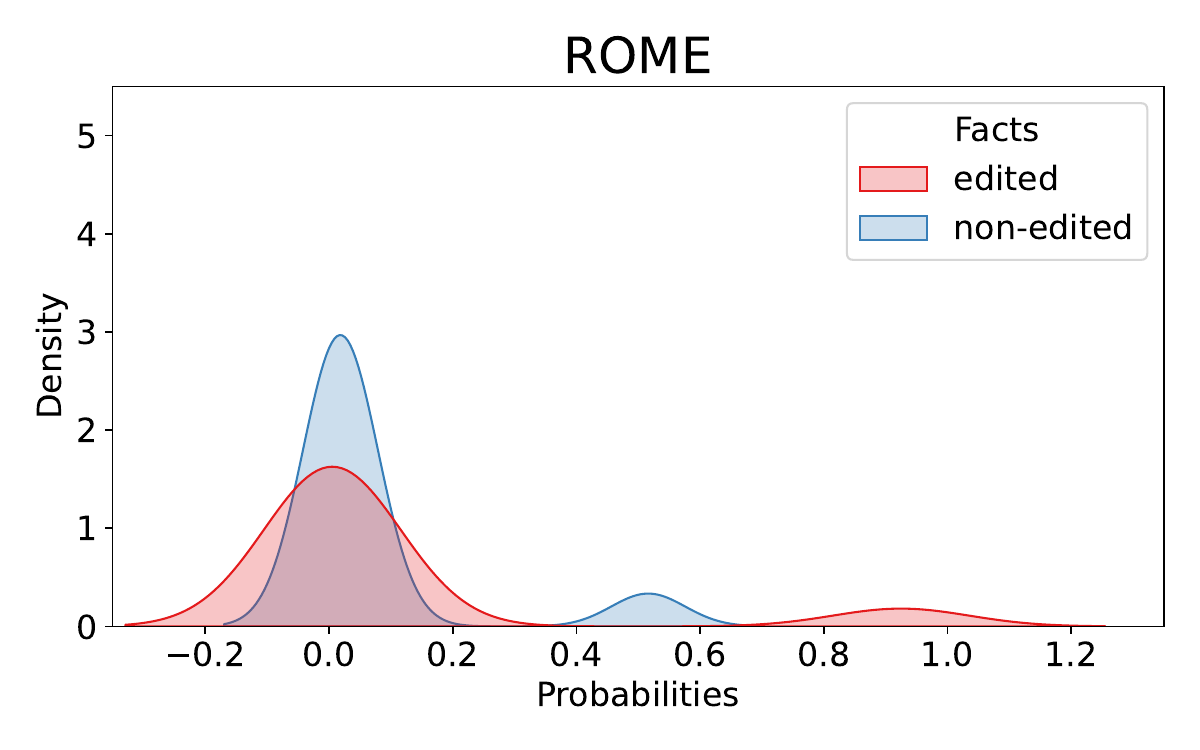}\\
  \includegraphics[width=\columnwidth, trim={0.5cm 0cm 0.5cm 0.3cm}, clip]{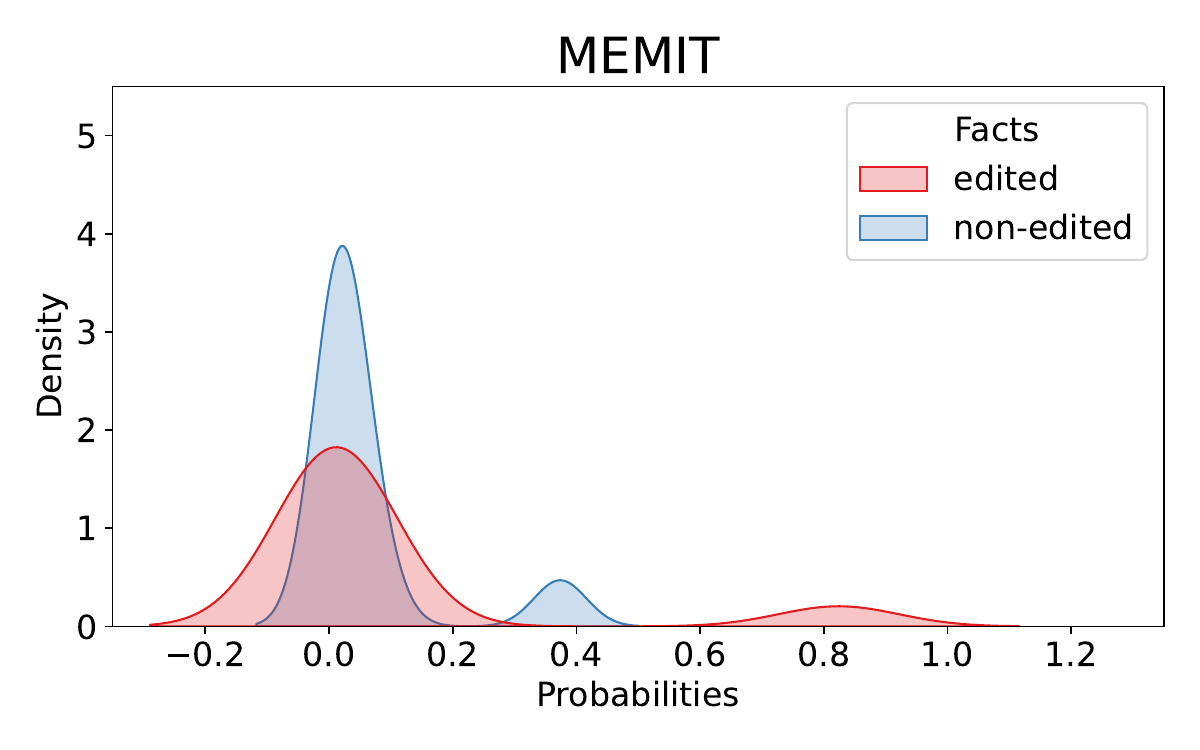}\\
  \includegraphics[width=\columnwidth, trim={0.5cm 0cm 0.5cm 0.3cm}, clip]{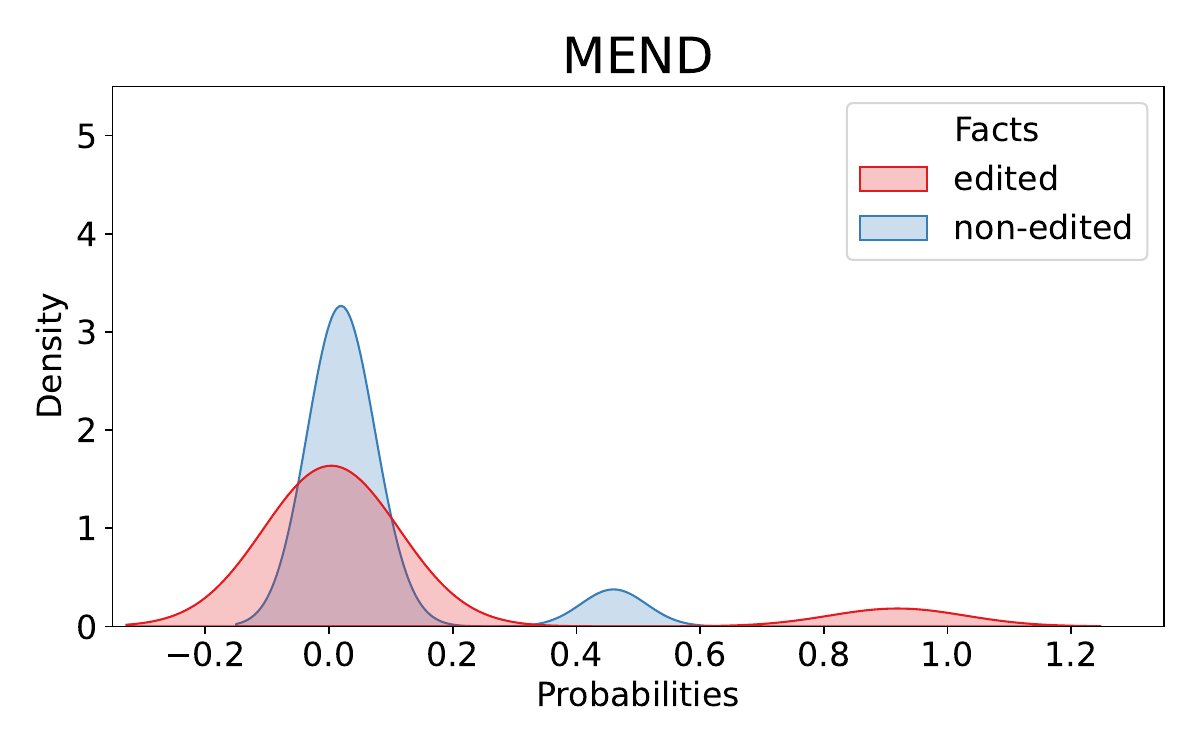}\\
  \includegraphics[width=\columnwidth, trim={0.5cm 0cm 0.5cm 0.3cm}, clip]{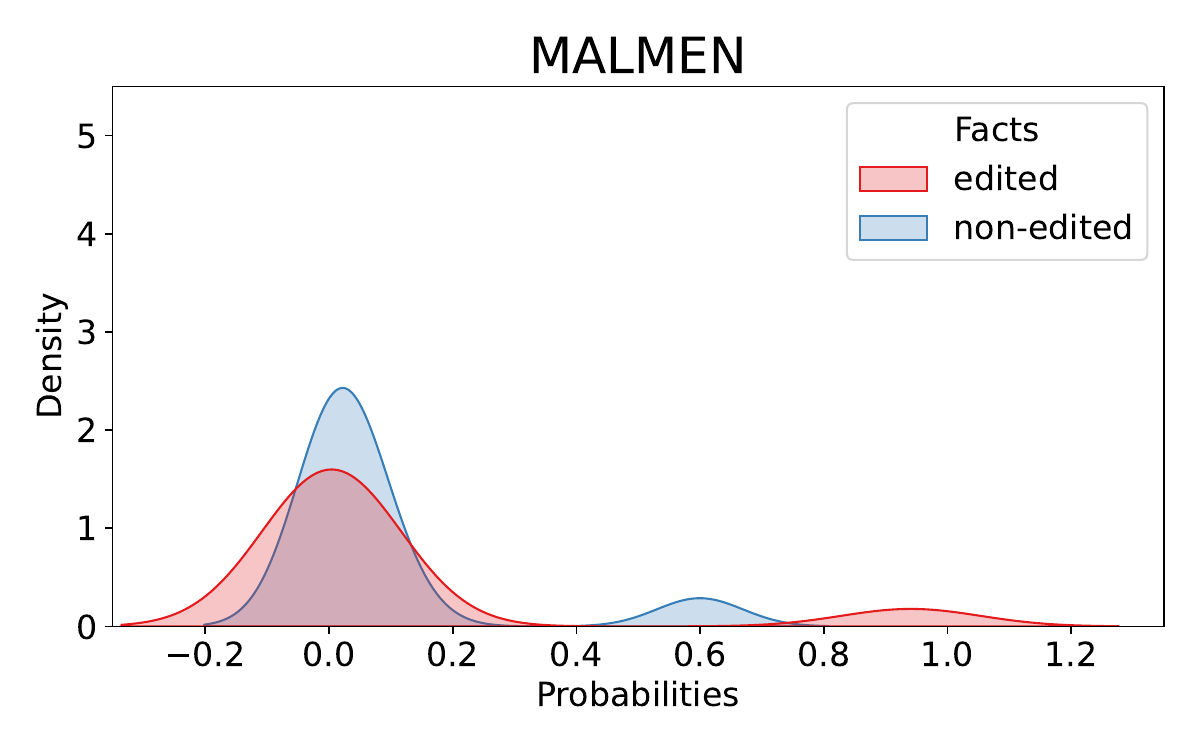}\\

\caption{\zsre and GPT-J.}
\label{fig:kde:cfd:gptj}
    \end{subfigure}
    \hfill
    \begin{subfigure}[t]{0.32\textwidth}
        \centering
  \includegraphics[width=\columnwidth, trim={0.5cm 0cm 0.5cm 0.3cm}, clip]{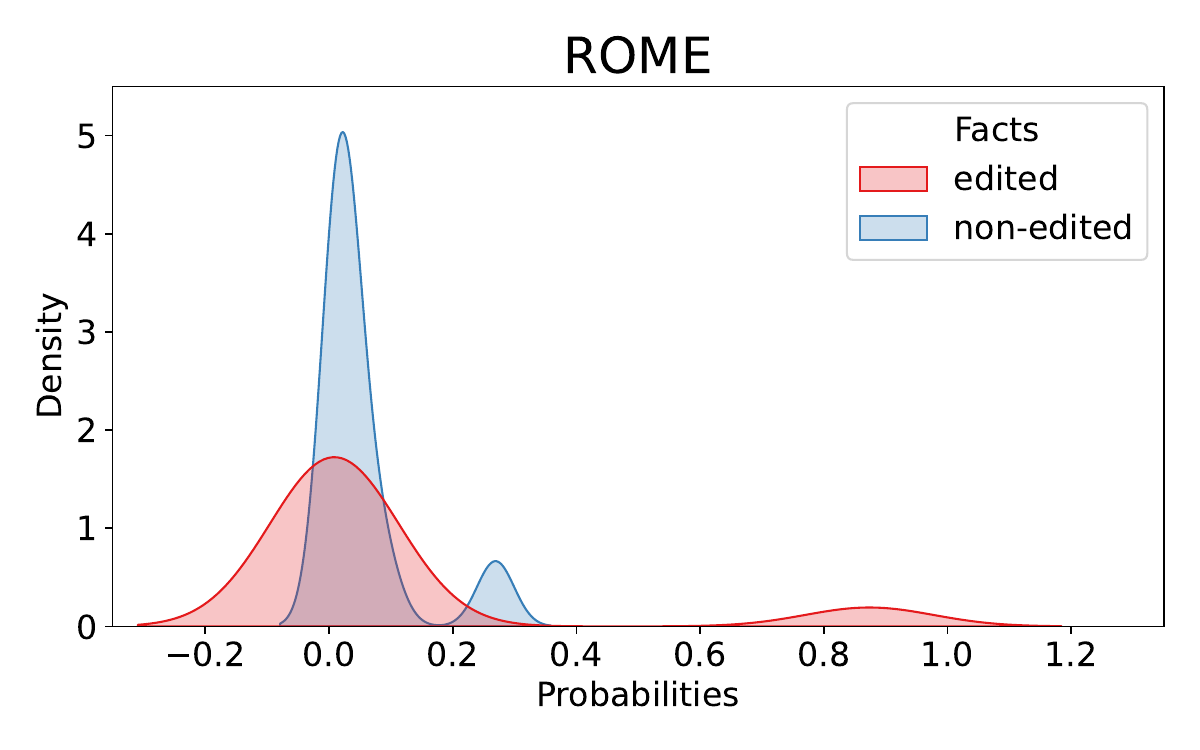}\\
  \includegraphics[width=\columnwidth, trim={0.5cm 0cm 0.5cm 0.3cm}, clip]{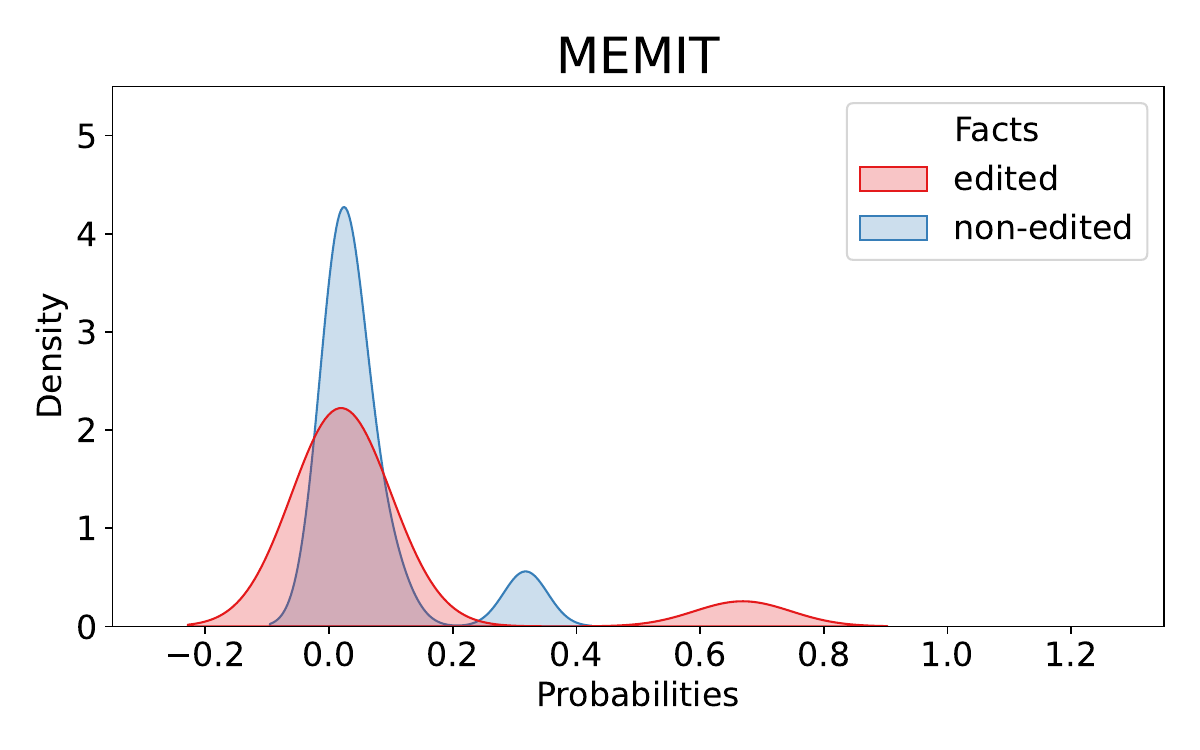}\\
  \includegraphics[width=\columnwidth, trim={0.5cm 0cm 0.5cm 0.3cm}, clip]{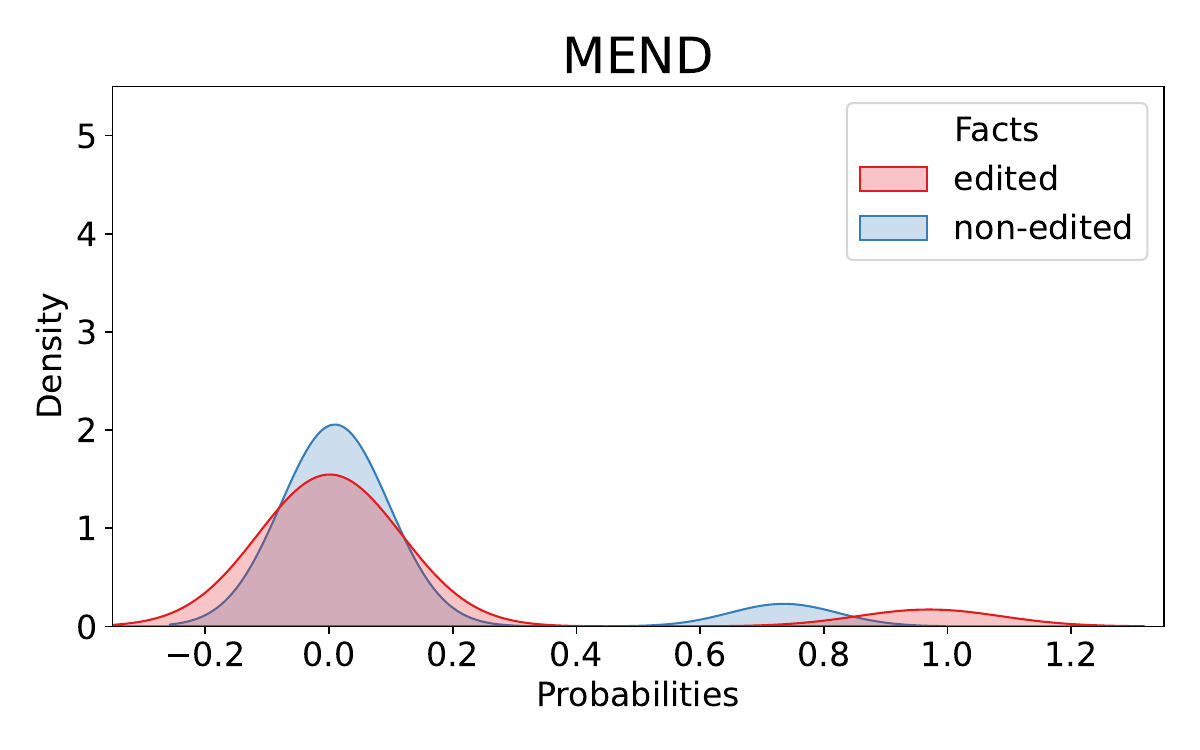}\\
  \includegraphics[width=\columnwidth, trim={0.5cm 0cm 0.5cm 0.3cm}, clip]{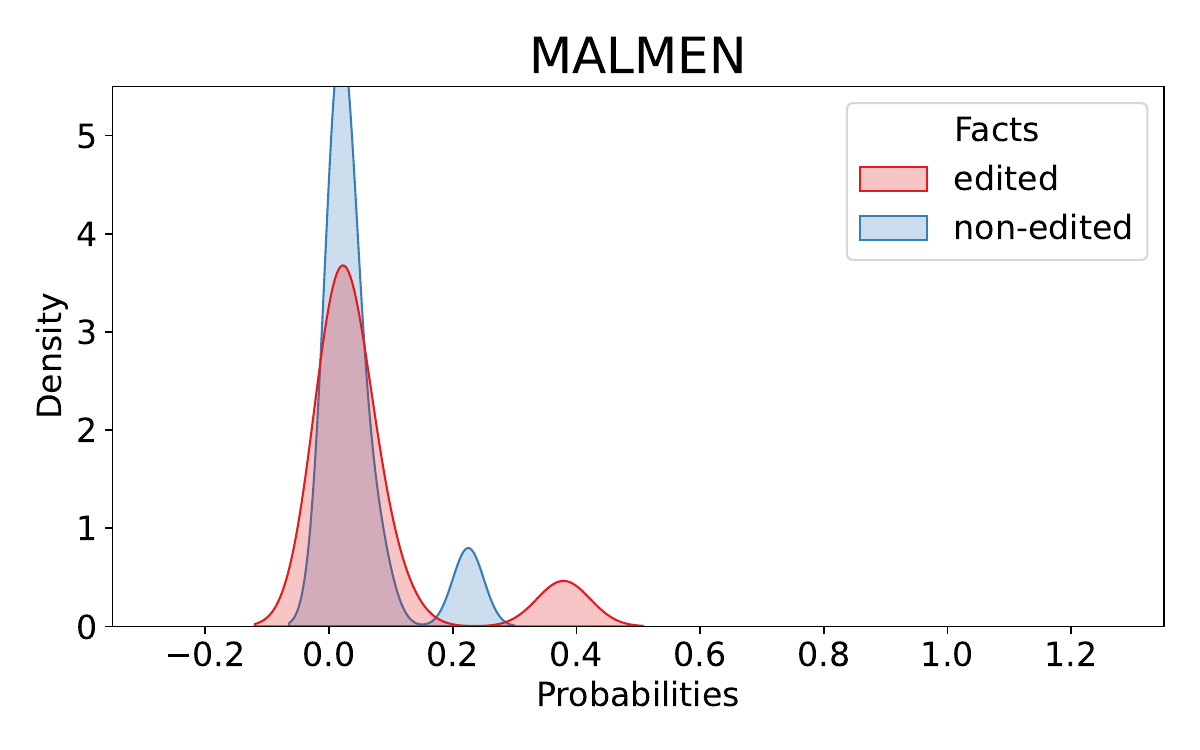}\\

\caption{\cfd and GPT2-XL.}
\label{fig:kde:cfd:gpt2}
    \end{subfigure}
    \hfill
    \begin{subfigure}[t]{0.32\textwidth}
        \centering
  \includegraphics[width=\columnwidth, trim={0.5cm 0cm 0.5cm 0.3cm}, clip]{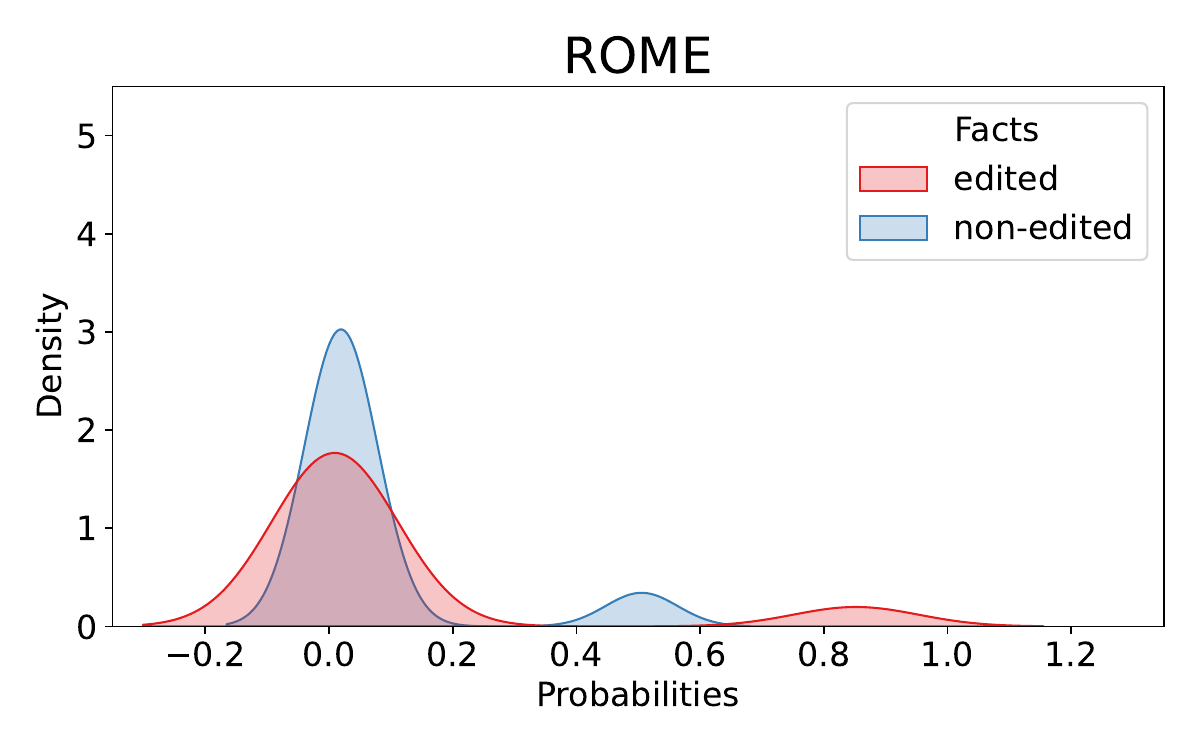}\\
  \includegraphics[width=\columnwidth, trim={0.5cm 0cm 0.5cm 0.3cm}, clip]{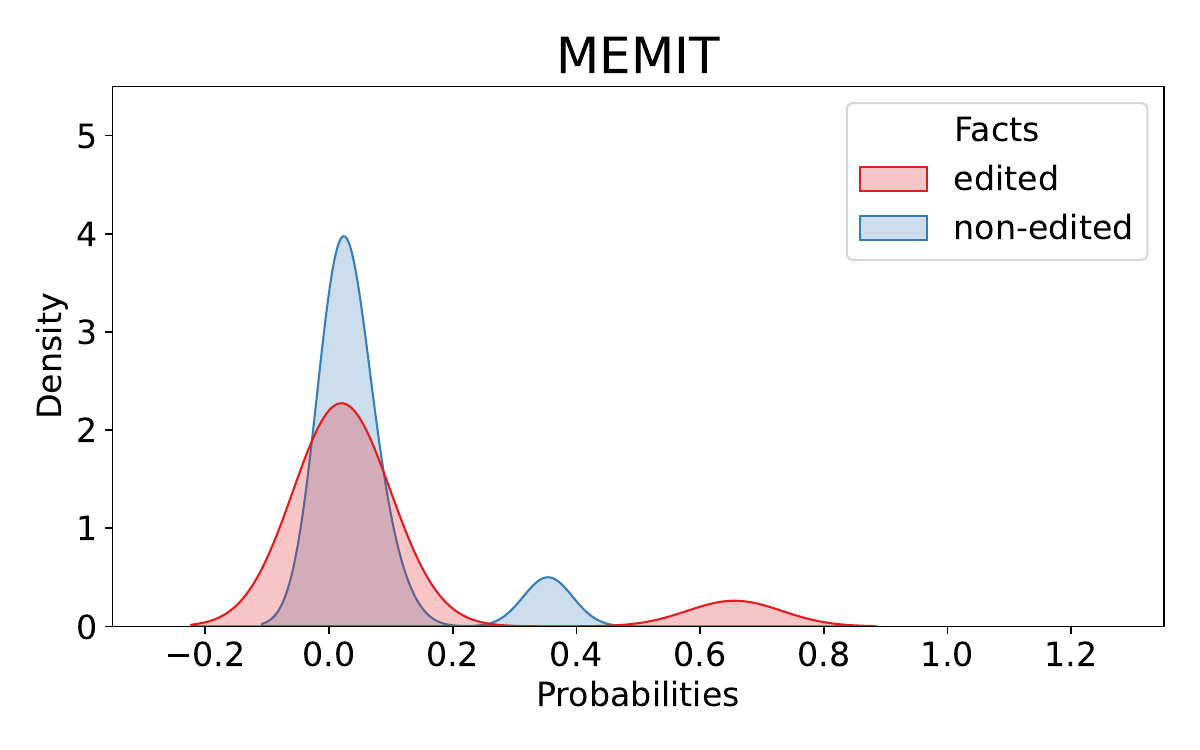}\\
  \includegraphics[width=\columnwidth, trim={0.5cm 0cm 0.5cm 0.3cm}, clip]{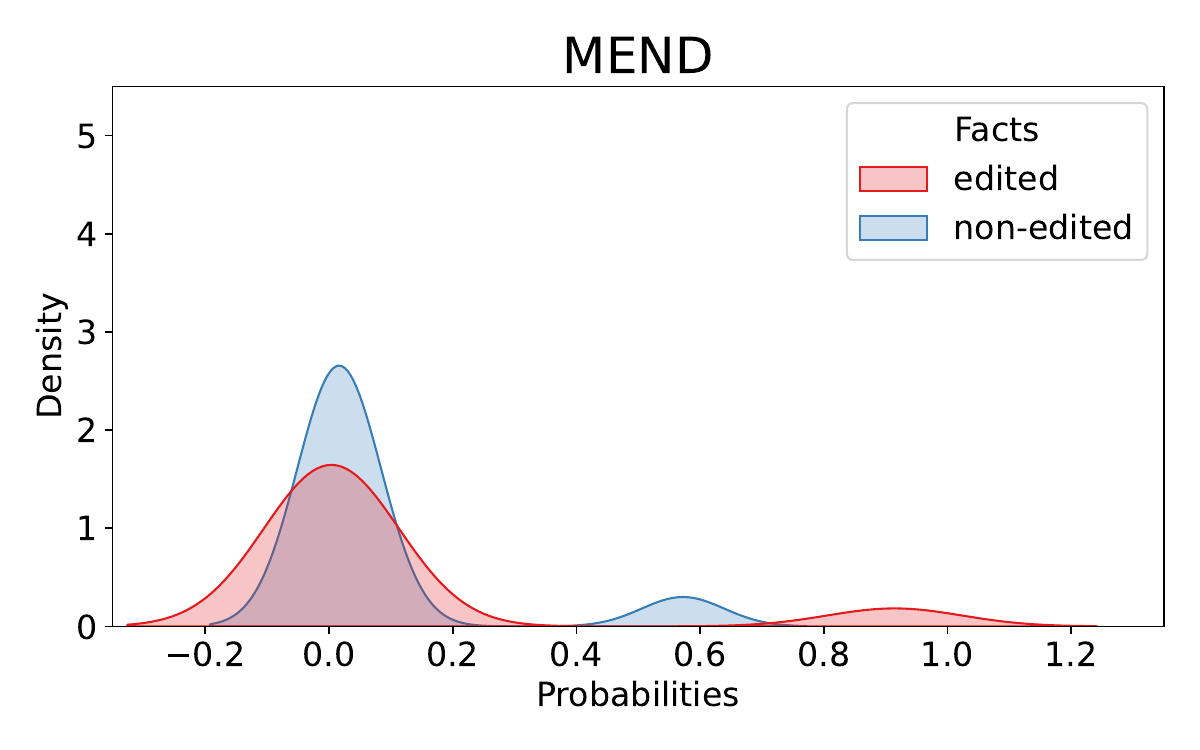}\\
  \includegraphics[width=\columnwidth, trim={0.5cm 0cm 0.5cm 0.3cm}, clip]{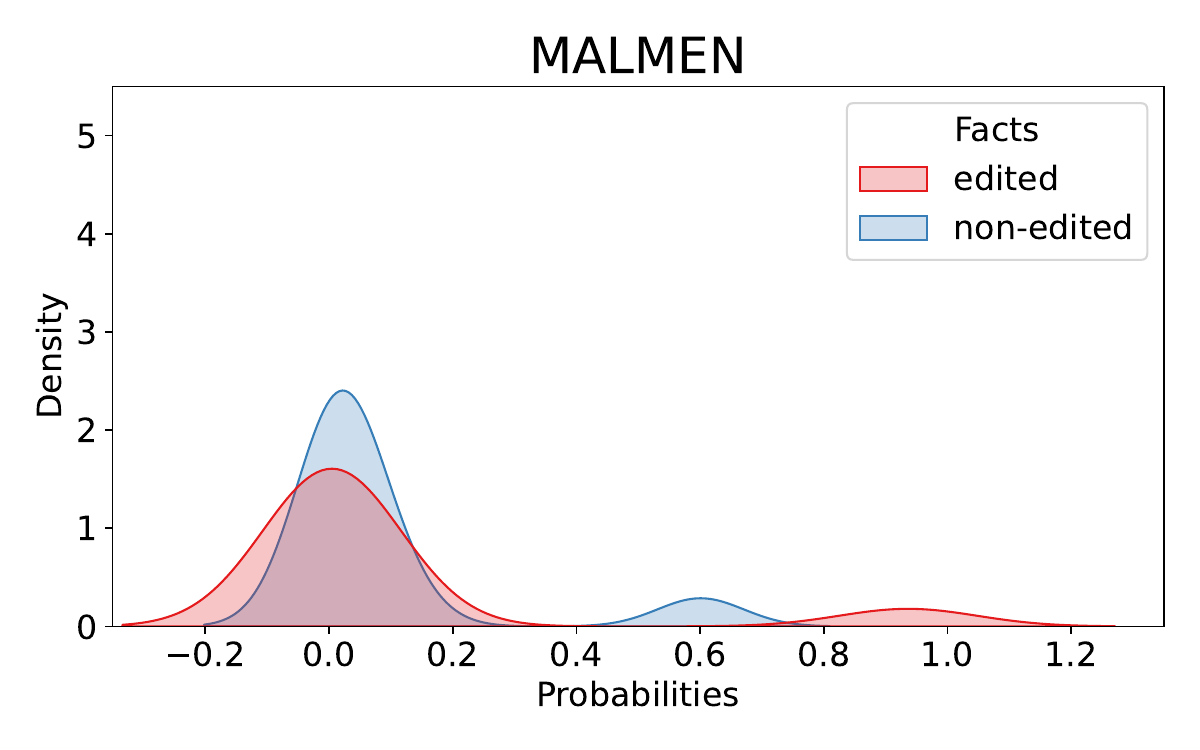}\\

\caption{\zsre and GPT2-XL.}

\label{fig:kde:zsre:gpt2}
    \end{subfigure}
    \caption{KDE of edited and non-edited facts.}
    \label{fig:app_kde}
\end{figure*}

In Table~\ref{tab:data_stats}, we report the number of training and test instances used in our default setting. We keep the number of training instances constant across all settings for consistency. Results for varying sizes of the training set are presented in Table~\ref{tab:cls_data} and Figure~\ref{fig:classification_performance_over_data:cfd} (\cfd). The result for the \crossdomain evaluation is shown in Table~\ref{tab:ood}. We show some prompt examples that are used for editing and evaluating edits in Table~\ref{tab:prompt_examples}. We show the \hs{} representations in a low-dimensional space in Figure~\ref{fig:app_lda}. We show KDE visualizations of \pd{} across all datasets, models, and KEs in Figure~\ref{fig:app_kde}. 

\label{app:full_edit_performance}
Editing performance scores are shown in Table~\ref{tab:edit_score}. Editing methods are typically evaluated in terms of Efficacy, Generalizability and Locality Success.
Efficacy Success (\textbf{ES}) is the primary evaluation measure and refers to the ability to change the LLM's predictions. 
For example, for the prompt $p(s,r)$ ``The Eiffel Tower is in the city of'' originally retrieving object $o$ ``Paris'', an edit is successful, if afterwards the desired object $o^{\prime}$ (e.g. ``Berlin'') is retrieved.
Ideally, such edits should also affect semantically similar prompts $p^{\prime}(s,r)$, e.g., ``The city where the Eiffel Tower is located is''. The ability of edits to generalize to similar prompts is referred to as \textbf{GS} (Generalizability Success). Edits should not affect other facts. \textbf{LS} (Locality Success, also known as Specificity)~\cite{pmlr-v162-mitchell22a,meng-etal-2022-locating, Li2024pmet} measures the accuracy of the model on a set of unrelated facts.

\begin{table}[h]
\centering
\resizebox{\columnwidth}{!}{%
\begin{tabular}{@{}llcccccc@{}}
\toprule
\multicolumn{2}{c}{Generator} & \multicolumn{3}{c}{\zsre} & \multicolumn{3}{c}{\cfd} \\ \midrule
Model & Editor & ES\textuparrow & GS\textuparrow & LS & ES\textuparrow & GS\textuparrow & LS\\ \midrule
\multirow{5}{*}{GPT-J} & NONE &   28.1 &   27.4 &   26.1 &   13.3 &   18.6 &   81.0 \\ \cmidrule(l){2-8} 
&   ROME &   99.5 &   94.4 &   25.5 &  100.0 &   99.8 &   79.4 \\
&  MEMIT &   98.7 &   90.6 &   35.4 &   99.4 &   90.8 &   78.8 \\
&   MEND &   92.6 &   89.0 &   20.3 &   91.4 &   79.5 &   70.4 \\
& MALMEN &   99.8 &   94.8 &   44.3 &   97.8 &   60.9 &   66.2 \\
 \midrule
\multirow{5}{*}{GPT2-XL} & NONE &   23.2 &   22.1 &   22.3 &   18.9 &   24.1 &   74.6 \\ \cmidrule(l){2-8} 
 &   ROME &   99.8 &   87.7 &   22.5 &   99.9 &   98.1 &   75.2 \\
 &  MEMIT &   68.8 &   58.1 &   27.6 &   93.7 &   81.1 &   74.8 \\
  &   MEND &   91.2 &   89.2 &   15.0 &   91.9 &   81.5 &   65.3 \\
 & MALMEN &   99.7 &   93.0 &   39.6 &   85.1 &   44.8 &   66.3 \\
 \bottomrule
\end{tabular}%
}
\caption{Summary of editing performance over 1000 edits. %
}
\label{tab:edit_score}
\end{table}

\begin{table}[h]
    \centering
    \resizebox{\columnwidth}{!}{%
\begin{tabular}{lllrr}
\toprule
   model &          ds & method &  \#training instances &  \#test instances \\
\midrule
gpt-j-6B & \cfd & MALMEN &                  126 &              468 \\
gpt-j-6B & \cfd &  MEMIT &                  126 &             1118 \\
gpt-j-6B & \cfd &   MEND &                  126 &             1390 \\
gpt-j-6B & \cfd &   NONE &                  126 &             1874 \\
gpt-j-6B & \cfd &   ROME &                  126 &             1548 \\
gpt-j-6B &        \zsre & MALMEN &                  126 &             1630 \\
gpt-j-6B &        \zsre &  MEMIT &                  126 &             1474 \\
gpt-j-6B &        \zsre &   MEND &                  126 &             1518 \\
gpt-j-6B &        \zsre &   NONE &                  126 &             1874 \\
gpt-j-6B &        \zsre &   ROME &                  126 &             1626 \\
 gpt2-xl & \cfd & MALMEN &                  126 &              128 \\
 gpt2-xl & \cfd &  MEMIT &                  126 &              708 \\
 gpt2-xl & \cfd &   MEND &                  126 &             1386 \\
 gpt2-xl & \cfd &   NONE &                  126 &             1874 \\
 gpt2-xl & \cfd &   ROME &                  126 &             1318 \\
 gpt2-xl &        \zsre & MALMEN &                  126 &             1538 \\
 gpt2-xl &        \zsre &  MEMIT &                  126 &              522 \\
 gpt2-xl &        \zsre &   MEND &                  126 &             1404 \\
 gpt2-xl &        \zsre &   NONE &                  126 &             1874 \\
 gpt2-xl &        \zsre &   ROME &                  126 &             1346 \\
    \bottomrule
    \end{tabular}%
}
    \caption{Number of training and test instances used in our default setting. Both the training and test sets are balanced (50\% for each class).}
    \label{tab:data_stats}
\end{table}

\begin{table*}[h!]
\centering
\resizebox{\textwidth}{!}{%
\begin{tabular}{ll@{\hskip 5mm}lll@{\hskip 7mm}lll@{\hskip 10mm}lll@{\hskip 7mm}lll@{\hskip 10mm}lll@{\hskip 7mm}lll}

\toprule

&  
& \multicolumn{6}{l}{\textbf{Hidden States + Prob. Dist. (HS + PD)}} 

& \multicolumn{6}{c}{\textbf{Hidden States (HS)}} 
& \multicolumn{6}{c}{\textbf{Probability Distribution (PD)}}\\ 
& 
& \multicolumn{3}{c}{\zsre} & \multicolumn{3}{l}{\cfd} 
& \multicolumn{3}{c}{\zsre} & \multicolumn{3}{l}{\cfd} 
& \multicolumn{3}{c}{\zsre} & \multicolumn{3}{l}{\cfd}\\ \midrule
 & KE &  Pr. &  Rec. &  F1   &  Pr. &  Rec. &  F1 &  Pr. &  Rec. &  F1   &  Pr. &  Rec. &  F1 &  Pr. &  Rec. &  F1   &  Pr. &  Rec. &  F1 \\ \midrule
\multirow{4}{*}{\rotatebox[origin=c]{90}{GPT-JT}} 

 &   NONE &          49.2 &       27.8 &   35.6 &          50.4 &       49.5 &   50.0  &          48.4 &       32.8 &   39.1 &          49.6 &       43.5 &   46.4 &            0.0 &        0.0 &    0.0 &          50.2 &       39.7 &   44.3 \\
 &   ROME &          87.1 &       95.5 &   91.1 &          90.9 &       94.2 &   92.5  &          99.8 &       74.2 &   85.1 &          91.6 &       87.8 &   89.6 &           84.8 &       84.5 &   84.7 &          89.3 &       94.0 &   91.6 \\
 &  MEMIT &          84.3 &       75.0 &   79.4 &          76.6 &       81.8 &   79.1  &          80.5 &       56.2 &   66.2 &          69.9 &       76.5 &   73.0 &           91.6 &       68.7 &   78.5 &          79.3 &       84.7 &   81.9 \\
 & MALMEN &          80.7 &       71.6 &   75.9 &          65.6 &       74.1 &   69.6  &          50.4 &       34.6 &   41.1 &          57.8 &       69.1 &   63.0 &           72.8 &       88.4 &   79.9 &          82.4 &       54.0 &   65.2 \\

 \midrule
\multirow{4}{*}{\rotatebox[origin=c]{90}{GPT2 lima}}

&   NONE &          49.0 &       56.1 &   52.3 &          51.5 &       45.9 &   48.5 &          50.1 &       41.8 &   45.6 &          49.4 &       42.0 &   45.4 & 50.6 &       93.0 &   65.6 &          51.2 &       39.2 &   44.4 \\
&   ROME &          91.4 &       99.0 &   95.0 &          89.6 &       93.8 &   91.6 &          97.8 &       93.2 &   95.4 &          94.5 &       83.0 &   88.4 & 63.6 &       86.9 &   73.4 &          88.7 &       91.3 &   90.0 \\
&  MEMIT &          79.2 &       65.0 &   71.4 &          77.3 &       77.1 &   77.2  &          76.9 &       61.9 &   68.6 &          73.7 &       60.9 &   66.7 & 74.0 &       73.2 &   73.6 &          76.9 &       76.4 &   76.7 \\
& MALMEN &          73.4 &       77.7 &   75.5 &          68.5 &       75.5 &   71.8 &          48.3 &       53.9 &   51.0 &          65.0 &       53.1 &   58.4 & 73.6 &       92.0 &   81.8 &          68.1 &       65.3 &   66.7 \\

 \bottomrule
\end{tabular}%
}
\caption{CD (Cross-Domain) classification performance. Training data is from GPT2-XL and GPT-J, whereas test data is from GPT2-XL lima (GPT2 lima) and GPT-JT respectively.}
\label{tab:ood}
\end{table*}

\begin{table*}[htp]
\scriptsize
    \centering
\begin{tabular}{clrcrrrrrrrrr}
\toprule
& & & & \multicolumn{4}{c}{\zsre} & & \multicolumn{4}{c}{\cfd} \\
\cmidrule{5-8}\cmidrule{10-13}
  model & method &  \#training & &  Acc. &  Pr. &  Rec. &  F1 & &  Acc. &  Pr. &  Rec. &  F1 \\
\midrule
  \multirow{34}{*}{\rotatebox{90}{GPT-J}} & MALMEN &                   10 & &    74.8 &          80.0 &       66.3 &   72.5 & &    51.3 &          50.7 &       94.6 &   66.0 \\
   & MALMEN &                   50 & &    73.7 &          73.8 &       73.4 &   73.6 & &    61.1 &          60.4 &       64.4 &   62.3 \\
   & MALMEN &                  100 & &    73.5 &          71.0 &       79.3 &   74.9 & &    64.4 &          63.9 &       66.4 &   65.1 \\
   & MALMEN &                  150 & &    75.3 &          74.9 &       76.1 &   75.5 & &    60.1 &          59.7 &       61.7 &   60.7 \\
   & MALMEN &                  200 & &    75.7 &          74.8 &       77.7 &   76.2 & &    65.8 &          65.6 &       66.4 &   66.0 \\
   & MALMEN &                  250 & &    76.1 &          76.3 &       75.6 &   76.0 & &    65.8 &          67.4 &       61.1 &   64.1 \\
  
   &  MEMIT &                   10 & &    84.6 &          78.9 &       94.5 &   86.0 & &    73.6 &          68.5 &       87.5 &   76.8 \\
   &  MEMIT &                   50 & &    83.8 &          77.6 &       95.0 &   85.4 & &    78.8 &          78.1 &       80.1 &   79.0 \\
   &  MEMIT &                  100 & &    86.5 &          84.0 &       90.2 &   87.0 & &    77.3 &          79.3 &       74.0 &   76.5 \\
   &  MEMIT &                  150 & &    87.0 &          86.1 &       88.2 &   87.2 & &    74.0 &          76.5 &       69.1 &   72.6 \\
   &  MEMIT &                  200 & &    86.4 &          86.6 &       86.0 &   86.3 & &    76.2 &          76.4 &       75.9 &   76.1 \\
   &  MEMIT &                  250 & &    86.8 &          85.2 &       89.0 &   87.0 & &    76.4 &          78.1 &       73.3 &   75.6 \\
   &  MEMIT &                  300 & &    86.8 &          84.5 &       90.0 &   87.2 & &    76.7 &          75.6 &       78.8 &   77.2 \\
  
   &   MEND &                   10 & &    72.1 &          79.2 &       60.1 &   68.3 & &    73.2 &          66.3 &       94.5 &   77.9 \\
   &   MEND &                   50 & &    83.2 &          79.6 &       89.3 &   84.2 & &    69.8 &          65.8 &       82.6 &   73.2 \\
   &   MEND &                  100 & &    79.2 &          80.3 &       77.4 &   78.8 & &    69.1 &          63.6 &       89.2 &   74.3 \\
   &   MEND &                  150 & &    81.0 &          77.4 &       87.6 &   82.2 & &    67.9 &          63.3 &       85.5 &   72.7 \\
   &   MEND &                  200 & &    81.6 &          78.5 &       87.1 &   82.6 & &    71.5 &          66.0 &       88.6 &   75.7 \\
   &   MEND &                  250 & &    82.1 &          79.2 &       87.1 &   83.0 & &    71.1 &          65.6 &       88.9 &   75.5 \\
   &   MEND &                  300 & &    81.3 &          78.0 &       87.1 &   82.3 & &    70.6 &          65.2 &       88.1 &   75.0 \\
  
   &   NONE &                   10 & &    47.9 &          47.0 &       33.2 &   38.9 & &    49.3 &          47.8 &       15.0 &   22.8 \\
   &   NONE &                   50 & &    51.1 &          51.2 &       48.0 &   49.5 & &    49.0 &          48.7 &       37.6 &   42.4 \\
   &   NONE &                  100 & &    50.5 &          50.6 &       43.4 &   46.7 & &    48.3 &          48.2 &       46.2 &   47.2 \\
   &   NONE &                  150 & &    47.1 &          46.7 &       41.6 &   44.0 & &    49.0 &          49.1 &       53.6 &   51.2 \\
   &   NONE &                  200 & &    48.2 &          48.1 &       46.6 &   47.4 & &    49.8 &          49.8 &       45.6 &   47.6 \\
   &   NONE &                  250 & &    47.8 &          47.5 &       42.2 &   44.7 & &    50.9 &          51.0 &       46.6 &   48.7 \\
   &   NONE &                  300 & &    49.3 &          49.3 &       52.2 &   50.7 & &    50.8 &          50.8 &       50.8 &   50.8 \\
  
   &   ROME &                   10 & &    94.8 &          99.5 &       90.0 &   94.5 & &    80.0 &          72.8 &       95.7 &   82.7 \\
   &   ROME &                   50 & &    95.6 &          97.4 &       93.6 &   95.5 & &    85.8 &          81.5 &       92.6 &   86.7 \\
   &   ROME &                  100 & &    96.6 &          97.4 &       95.7 &   96.5 & &    87.8 &          88.4 &       87.1 &   87.7 \\
   &   ROME &                  150 & &    96.7 &          98.1 &       95.2 &   96.6 & &    88.2 &          90.4 &       85.4 &   87.8 \\
   &   ROME &                  200 & &    96.5 &          97.7 &       95.2 &   96.4 & &    88.3 &          88.7 &       87.8 &   88.2 \\
   &   ROME &                  250 & &    96.0 &          97.6 &       94.3 &   95.9 & &    88.3 &          89.2 &       87.1 &   88.2 \\
   &   ROME &                  300 & &    96.2 &          97.6 &       94.8 &   96.2 & &    89.5 &          91.7 &       86.9 &   89.2 \\
  \midrule
 \multirow{31}{*}{\rotatebox{90}{GPT2-XL}}& MALMEN &                   10 & &    74.0 &          77.9 &       67.1 &   72.1 & &    46.1 &          46.5 &       51.6 &   48.9 \\
 & MALMEN &                   50 & &    71.5 &          65.6 &       90.6 &   76.1 & &    62.5 &          62.5 &       62.5 &   62.5 \\
 & MALMEN &                  100 & &    71.8 &          68.4 &       81.0 &   74.2 & &    65.6 &          65.6 &       65.6 &   65.6 \\
  
 &  MEMIT &                   10 & &    51.2 &          50.8 &       77.8 &   61.5 & &    58.8 &          63.1 &       42.6 &   50.9 \\
 &  MEMIT &                   50 & &    71.9 &          77.5 &       61.7 &   68.7 & &    60.8 &          62.0 &       55.5 &   58.6 \\
 &  MEMIT &                  100 & &    67.9 &          67.7 &       68.5 &   68.1 & &    69.6 &          72.3 &       63.6 &   67.7 \\
 &  MEMIT &                  150 & &    72.2 &          72.8 &       71.0 &   71.9 & &    73.2 &          73.9 &       71.8 &   72.8 \\
 &  MEMIT &                  200 & &    72.2 &          72.8 &       71.0 &   71.9 & &    73.7 &          75.4 &       70.3 &   72.8 \\
 &  MEMIT &                  250 & &    72.2 &          72.0 &       72.8 &   72.4 & &    72.0 &          72.1 &       71.8 &   71.9 \\
 &  MEMIT &                  300 & &    71.3 &          72.0 &       69.8 &   70.8 & &    72.2 &          74.4 &       67.9 &   71.0 \\
  
 &   MEND &                   10 & &    60.6 &          59.6 &       65.8 &   62.5 & &    50.1 &          50.4 &       16.9 &   25.4 \\
 &   MEND &                   50 & &    70.0 &          66.1 &       82.0 &   73.2 & &    60.8 &          58.6 &       73.8 &   65.3 \\
 &   MEND &                  100 & &    69.7 &          66.1 &       80.9 &   72.8 & &    59.9 &          56.8 &       82.5 &   67.3 \\
 &   MEND &                  150 & &    70.2 &          66.0 &       83.6 &   73.7 & &    64.7 &          59.0 &       95.8 &   73.1 \\
 &   MEND &                  200 & &    71.8 &          67.7 &       83.3 &   74.7 & &    65.0 &          59.3 &       95.2 &   73.1 \\
 &   MEND &                  250 & &    72.8 &          68.0 &       86.2 &   76.0 & &    65.1 &          59.4 &       95.5 &   73.2 \\
 &   MEND &                  300 & &    73.0 &          68.1 &       86.4 &   76.2 & &    65.1 &          59.4 &       95.5 &   73.2 \\
  
 &   NONE &                   10 & &    50.2 &          50.6 &       17.0 &   25.4 & &    50.5 &          50.8 &       31.2 &   38.7 \\
 &   NONE &                   50 & &    52.0 &          52.3 &       45.6 &   48.7 & &    50.8 &          51.0 &       41.6 &   45.8 \\
 &   NONE &                  100 & &    50.1 &          50.1 &       46.8 &   48.4 & &    48.7 &          48.7 &       49.8 &   49.3 \\
 &   NONE &                  150 & &    52.0 &          52.4 &       44.6 &   48.2 & &    46.9 &          46.6 &       43.2 &   44.9 \\
 &   NONE &                  200 & &    49.8 &          49.8 &       46.4 &   48.0 & &    48.6 &          48.6 &       50.6 &   49.6 \\
 &   NONE &                  250 & &    49.4 &          49.3 &       44.2 &   46.6 & &    48.8 &          48.8 &       49.4 &   49.1 \\
 &   NONE &                  300 & &    52.8 &          52.5 &       58.4 &   55.3 & &    49.3 &          49.4 &       54.8 &   51.9 \\
  
 &   ROME &                   10 & &    88.2 &          88.5 &       87.8 &   88.1 & &    63.6 &          58.4 &       93.9 &   72.0 \\
 &   ROME &                   50 & &    91.2 &          95.8 &       86.1 &   90.7 & &    87.1 &          89.4 &       84.2 &   86.7 \\
 &   ROME &                  100 & &    91.2 &          92.4 &       89.7 &   91.0 & &    89.2 &          91.7 &       86.2 &   88.9 \\
 &   ROME &                  150 & &    92.5 &          95.6 &       89.1 &   92.3 & &    86.7 &          85.5 &       88.4 &   86.9 \\
 &   ROME &                  200 & &    92.0 &          94.5 &       89.1 &   91.8 & &    88.0 &          89.4 &       86.2 &   87.7 \\
 &   ROME &                  250 & &    90.9 &          92.6 &       88.9 &   90.7 & &    88.8 &          90.0 &       87.3 &   88.6 \\
 &   ROME &                  300 & &    90.8 &          93.4 &       87.8 &   90.5 & &    88.6 &          90.0 &       87.0 &   88.4 \\
\bottomrule
\end{tabular}
    \caption{DEED (\textbf{de}tecting knowledge \textbf{ed}its) performance with increasing amount of training data.}
    \label{tab:cls_data}
\end{table*}

\subsection{HS from Different Layers}
\label{app:hs_layers}
\begin{figure*}[h]
\centering
  \includegraphics[width=0.95\textwidth, trim={0.3cm 0.1cm 2.5cm 0cm}, clip]{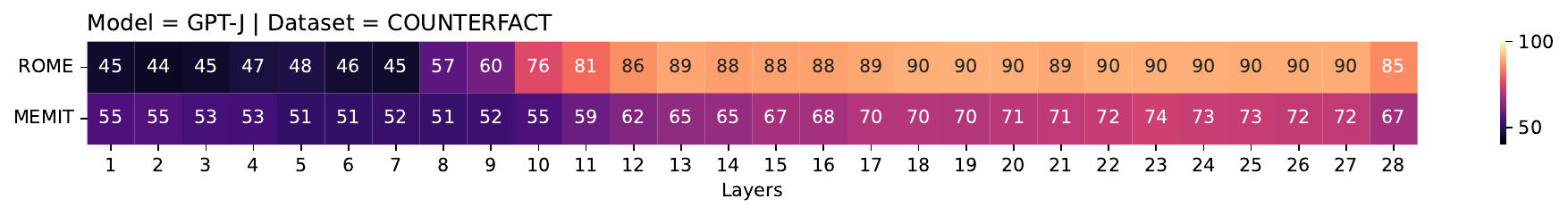}\\
  \includegraphics[width=0.95\textwidth, trim={0.3cm 0.1cm 2.5cm 0cm}, clip]{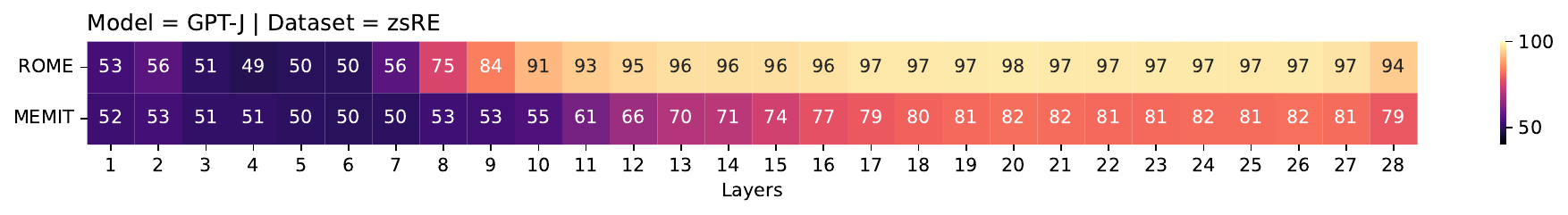}\\
  \includegraphics[width=0.95\textwidth, trim={0.3cm 0.1cm 2.5cm 0cm}, clip]{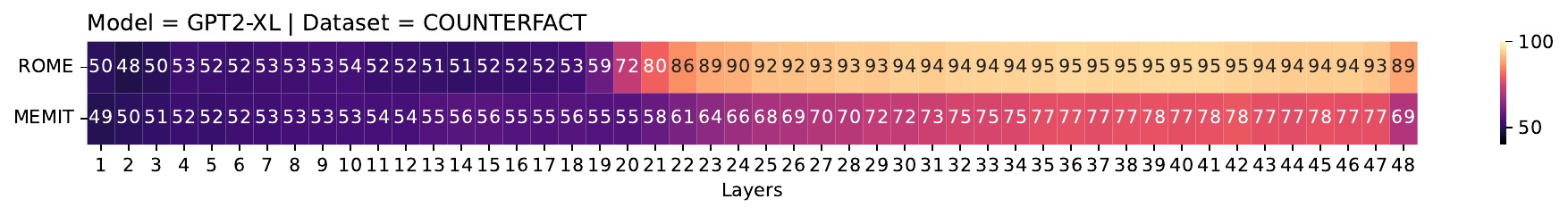}\\
  \includegraphics[width=0.95\textwidth, trim={0.3cm 0.1cm 2.5cm 0cm}, clip]{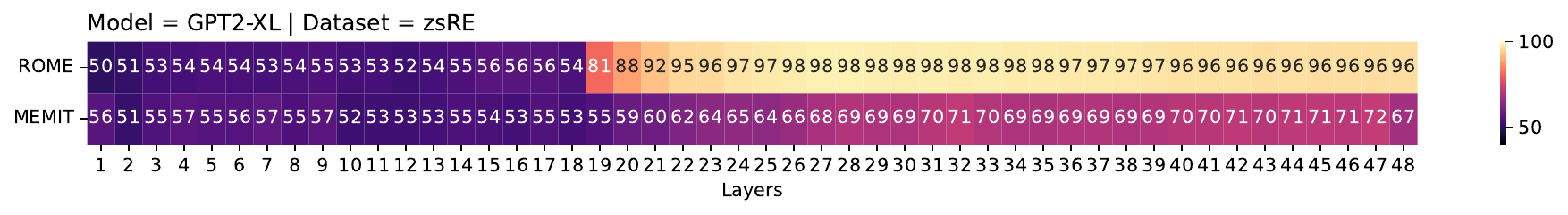}
\caption{Rounded F1 scores when training on representations (\hs{}) from different LLM layers}%
\label{fig:heatmap}
\end{figure*}

To investigate the effect of \hs{} from different layers on the classification performance, we train a simple logistic regression classifier with L1 regularization on \hs from different layers. We consider only locate-and-edit methods (ROME and MEMIT) for this experiment, since \hs{} performs well on this category of methods. We show the F1 values based on hidden states representations from different layers in Figure~\ref{fig:heatmap}. 
First, for both models, editing methods, and datasets, the representations start becoming separable, i.e., the classification performance is above random guess, after the edited layers or shortly thereafter. For example, the F1 for detecting ROME edits starts improving from the layer 7-8 in GPT-J and the layer 19 in GPT2-XL (ROME changes layer 7 in GPT-J, and layer 19 in GPT2-XL). We observe the same effect with MEMIT (MEMIT changes the layers 5-10 in GPT-J and the layers 15-19 in GPT2-XL). %
Even though the representations from the last layer perform well in detecting edits, we notice that the F1 reaches its peak at preceding layers. For ROME, the highest performance can be seen at training on representations from the layer 13-27 (GPT-J) or from the layers 23-47 (GPT2-XL). For MEMIT, the highest F1 is reached with representations from layer 17-27 (GPT-J), and layer 28-47 (GPT2-XL). This suggests that utilizing outputs from the last layer for detecting knowledge edits might not always be the optimal choice. Further, we observe that using the penultimate layer's representations is more effective than using the last layer's to detect knowledge edits. For example, for edited GPT-J on \cfd (first subplot in Figure \ref{fig:heatmap}), representations from the penultimate layer achieve an F1 of 90\% and 72\% for detecting ROME-edited and MEMIT-edited knowledge, respectively, higher than 85\% (ROME) and 67\% (MEMIT) when using the representation from the last layer. \emph{LE-KEs might be causing the attribute (object) extraction phase~\cite{geva-etal-2023-dissecting} that is responsible for extracting the object to happen early in the model. This early extraction is often associated with high probabilities (cf. Section~\ref{subsec:analysis}), causing edits from LE-KEs to be easier to detect. In general, this experiment shows that the effect of edits from LE-KEs starts becoming visible from the changed layers, increases strongly thereafter, and decreases slightly at the last layer. This demonstrates that using representations from the last layer is not the optimal choice; utilizing representations from other layers can further improve performance}. %

\subsection{Robustness Evaluation}

To investigate the robustness of the classifier, which uses HS and PD, we further evaluate it in a challenging setting, where both edited and unedited facts end with the same object, i.e., the classifier needs to distinguish between edited = $(s_1, r_1, o', p'_1)$ and unedited = $(s_2, r_2, o', p'_2)$ facts that share the same object (e.g., ``The Eiffel Tower is in \underline{Berlin}'' vs. ``Marlene Dietrich was born in \underline{Berlin}''). 
Having the same object (as a prediction) in both edited and unedited facts might lead to more similar representations and probability distributions, and hence make the task more challenging. For example, the classifier might erroneously predict a fact as edited due to the presence of an edited fact with the same object.
We refer to this setting as \texttt{same object}. %
Concretely, the classification model and the training are the same as described in Section \ref{subsec:deed_dataset}. The test set is different. For each edited fact in the test set, there is a corresponding unedited fact with the same object. For this experiment, we consider only GPT-J with MEMIT and MALMEN.
The results for this experiment are shown in Table~\ref{tab:classification_performance:sameobject} alongside the corresponding results from Table~\ref{tab:classification_performance} (referred to as \texttt{default}) for comparison. We notice that the classification performance drops. The decrease in performance is especially visible on \zsre. For example, the F1 score with MEMIT and GPT-J drops from 88.7\% to 71.3\%, whereas on \cfd the F1 score drops from 77.9\% to 72.8\%. This experiment shows that \emph{it is more challenging to separate edited facts from unedited but related facts}. We believe this is due to the representations of the prompts becoming more similar under this setting.  %

\begin{table}[h!]
\centering
\resizebox{\columnwidth}{!}{%
\begin{tabular}{@{}llcccccc@{}}
\toprule
\multicolumn{2}{c}{Generator} & \multicolumn{3}{c}{\zsre} & \multicolumn{3}{c}{\cfd} \\ \midrule
Model & Editor &  Pr. &  Rec. &  F1   &  Pr. &  Rec. &  F1 \\ \midrule

\multirow{2}{*}{\texttt{default}} 
&  MEMIT &           87.7 &      89.8 &  88.7 &           79.0 &      76.7 &  77.9 \\

 & MALMEN &            76.7 &      75.3 &  76.0 &           62.0 &      62.0 &  62.0 \\ \midrule

\multirow{2}{*}{\texttt{same object}} 

&  MEMIT &          58.5 &       91.1 &   71.3 &          72.7 &       72.9 &   72.8 \\
& MALMEN &          57.5 &       57.9 &   57.7 &          56.4 &       63.6 &   59.8 \\

\bottomrule

\end{tabular}%
}
\caption{Classification performance on detecting knowledge edits for GPT-J. Facts in the training and test data end with the \texttt{same object}. Classification is done based on \hs{} and \pd{}.}
\label{tab:classification_performance:sameobject}
\end{table}

\subsection{Detection by Top 10 Output Probabilities}
We experiment with using only the top 10 output probabilities instead of the top 1,000 output probabilities. The differences in performance (top 10 performance - top 1000 performance)) in Table~\ref{tab:top10} show that the top-10 output probabilities works well with minor degradation in performance in most cases. This means that our baselines can also be used on proprietary LLMs that output the top 10 output probabilities.

\begin{table}[h!]
\centering
\resizebox{\columnwidth}{!}{%
\begin{tabular}{@{}llcccccc@{}}
\toprule
\multicolumn{2}{c}{Generator} & \multicolumn{3}{c}{\zsre} & \multicolumn{3}{c}{\cfd} \\ \midrule
Model & Editor &  Pr. &  Rec. &  F1   &  Pr. &  Rec. &  F1 \\ \midrule

\multirow{5}{*}{GPT-J} 
&   NONE &          -2.6 &        6.7 &    2.4 &          -2.4 &       -4.5 &   -3.5 \\
&   ROME &          -0.4 &       -0.4 &   -0.4 &           1.5 &       -1.5 &    0.0 \\
&  MEMIT &          -0.6 &       -0.9 &   -0.7 &          -1.7 &        2.6 &    0.6 \\
&   MEND &          -0.2 &        3.2 &    1.3 &           0.6 &       -0.4 &    0.3 \\
& MALMEN &           2.6 &       -1.9 &    0.5 &           4.0 &       -0.5 &    1.6 \\ \midrule
\multirow{5}{*}{GPT2-XL} 
 &   NONE &          -0.1 &        0.3 &    0.1 &          -0.8 &        0.0 &   -0.4 \\
 &   ROME &           0.1 &       -2.7 &   -1.3 &           2.0 &       -5.2 &   -1.5 \\
 &  MEMIT &          -1.4 &       -5.7 &   -3.8 &          -7.0 &       -1.7 &   -4.4 \\
 &   MEND &          -1.0 &       -5.8 &   -2.9 &           0.1 &        0.1 &    0.1 \\
 & MALMEN &           0.0 &        9.2 &    4.7 &          -4.6 &      -14.1 &   -9.4 \\

\bottomrule

\end{tabular}%
}
\caption{Differences in classification performance for detecting knowledge edits: top 10 probabilities minus top 1,000 probabilities}
\label{tab:top10}
\end{table}

\begin{table*}[h!]
\resizebox{\textwidth}{!}{%
\begin{tabular}{@{}lll@{}}
\toprule
Prompt example     & \zsre & \cfd                                                                              \\ \midrule
Edit prompt        & In what capacity did Andrea Guatelli play football? goakeeper   & Ramon Magsaysay holds a citizenship from Sweden                                      \\
Evaluate ES prompt &  In what capacity did Andrea Guatelli play football? goakeeper    & Ramon Magsaysay holds a citizenship from Sweden                                       \\
Evaluate GS prompt &   In what capacity has Andrea Guatelli played soccer? goalkeeper  & Marquez Rubio, Juan Carlos (2002). Ramon Magsaysay is a citizen of Sweden \\
Evaluate LS prompt &  What is the university Leslie Lazarus went to? University of Sydney   & Some fish, like sharks and lampreys, possess multiple gill openings. Abune Paulos, who is a citizen of Ethiopia                                                 \\ \bottomrule
\end{tabular}%
}
\caption{Example prompt used to edit knowledge, and to evaluate editing performance. Objects underlined are the original knowledge.}
\label{tab:prompt_examples}
\end{table*}

\section{Computational Resources}
We ran our experiments on an NVIDIA A100 80GB GPU. Our experiments took roughly 15 GPU days.

\section{Implementation Details}
We use EasyEdit~\cite{wang-etal-2024-easyedit} to edit with ROME, MEMIT and MEND. For MALMEN, we use the original implementation from the paper. 

\end{document}